\documentclass{article}

\usepackage[firstpage]{draftwatermark}
\SetWatermarkText{
 \hspace*{3.5in}
 \raisebox{10in}{
  \includegraphics[height=0.9in]{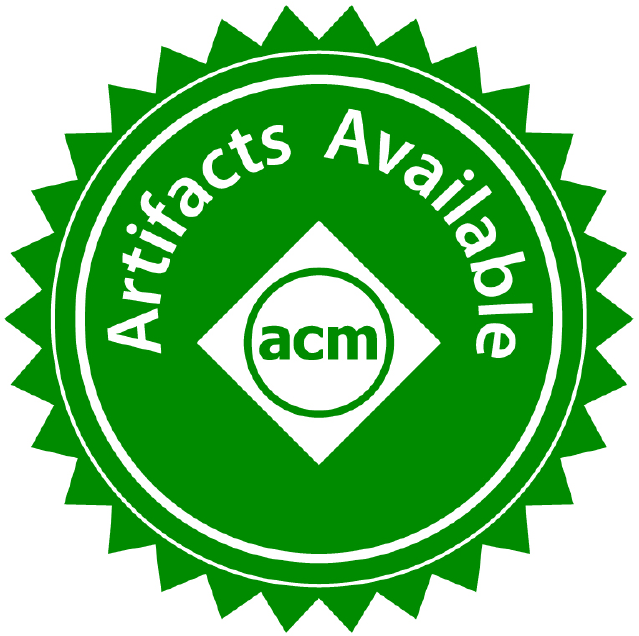}
  \includegraphics[height=0.9in]{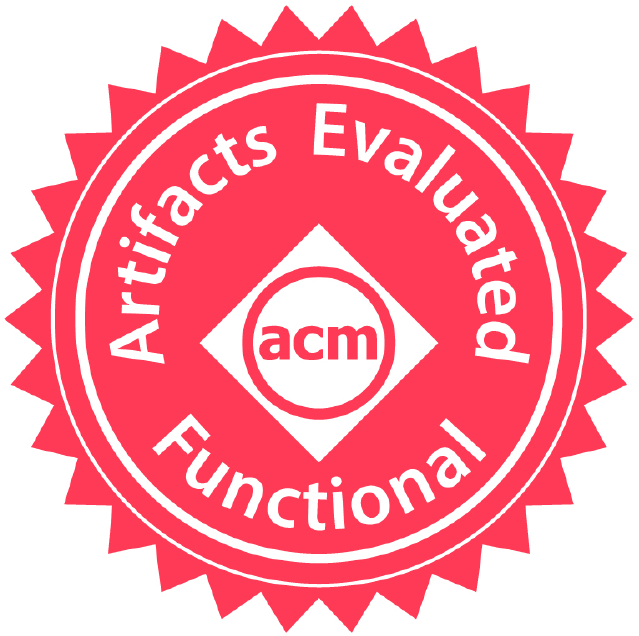}
 }
}
\SetWatermarkAngle{0}

\usepackage{microtype}
\usepackage{graphicx}
\usepackage{subfigure}
\usepackage{booktabs} 

\usepackage{hyperref}

\usepackage{amssymb}
\usepackage{amsmath}
\usepackage{amsthm}
\usepackage{multirow}

\newtheorem{theorem}{Theorem}
\newtheorem{assumption}{Assumption}

\usepackage[accepted]{mlsys2023}

\mlsystitlerunning{Adaptive Message Quantization and Parallelization for Distributed Full-graph GNN Training}

\begin{document}

\twocolumn[
\mlsystitle{Adaptive Message Quantization and Parallelization for
Distributed Full-graph GNN Training}



\mlsyssetsymbol{equal}{*}

\begin{mlsysauthorlist}
\mlsysauthor{Borui Wan}{hku}
\mlsysauthor{Juntao Zhao}{hku}
\mlsysauthor{Chuan Wu}{hku}
\end{mlsysauthorlist}

\mlsysaffiliation{hku}{Department of Computer Science, University of Hong Kong, Hong Kong, China}

\mlsyscorrespondingauthor{Borui Wan}{wanborui@connect.hku.hk}

\mlsyskeywords{Machine Learning, MLSys}

\vskip 0.3in

\begin{abstract}
Distributed full-graph training of Graph Neural Networks (GNNs) over large graphs is bandwidth-demanding and time-consuming. Frequent exchanges of node features, embeddings and embedding gradients (all referred to as \textit{messages}) across devices bring significant communication overhead for nodes with remote neighbors on other devices (\textit{marginal nodes}) and unnecessary waiting time for nodes without remote neighbors (\textit{central nodes}) in the graph. This paper proposes an efficient GNN training system, AdaQP, to expedite distributed full-graph GNN training. We stochastically quantize messages transferred across devices to lower-precision integers for communication traffic reduction and advocate communication-computation parallelization between marginal nodes and central nodes. We provide theoretical analysis to prove fast training convergence (at the rate of $O(T^{-1})$ with $T$ being the total number of training epochs) and design an adaptive quantization bit-width assignment scheme for each message based on the analysis, targeting a good trade-off between training convergence and efficiency. Extensive experiments on mainstream graph datasets show that AdaQP substantially improves 
distributed full-graph training's throughput (up to 3.01$\times$) with negligible accuracy drop (at most 0.30\%) or even accuracy improvement (up to 0.19\%) in most cases, showing significant advantages over the state-of-the-art works. The code is available at \href{https://github.com/raywan-110/AdaQP}{https://github.com/raywan-110/AdaQP}.
\end{abstract}
]


\printAffiliationsAndNotice{}  
\section{Introduction}
\label{sec:intro}
Graph Neural Networks (GNNs) have received increased attention from the AI community for their superior performance on graph-based tasks such as node classification~\cite{Kipf2017SemiSupervisedCW}, link prediction~\cite{Zhang2018LinkPB} and graph classification~\cite{Xu2019HowPA}. For each node of a graph, a GNN typically aggregates features or embeddings of the node's neighbors iteratively and then uses them to create the node's own embedding. This process is referred to as {\em message-passing}~\cite{Gilmer2017NeuralMP}, which enables GNNs to learn better representative embeddings from graph structures than traditional graph learning methods~\cite{wu2020comprehensive,zhang2020deep}.

For a $k$-layer GNN, the message-passing paradigm requires features and embeddings in the $k$-hop neighborhood of the training nodes to be retrieved and stored on a device (e.g., GPU) for computation, leading to high memory overhead. 
When training on large graphs, the memory consumption may easily exceed the memory capacity of a single device, and GNN training with graph sampling has hence been widely studied \cite{Hamilton2017InductiveRL, Chen2018FastGCNFL, Chiang2019ClusterGCNAE, Zeng2020GraphSAINTGS, wan2022bns}: the large input graphs are partitioned among multiple devices and machines; each worker (device) samples partial neighborhood of its training nodes and fetches features of sampled neighbors from other devices/machines if they are not in the local graph partition. 
Such graph sampling reduces memory, computation and communication overheads during distributed GNN training, at the cost of indispensable information loss for graph learning, as compared to full-graph training. Besides, sampling 
introduces extra time overhead due to 
running (sophisticated) sampling algorithms
~\cite{liu2021bgl, kaler2022accelerating}. 

Unlike sampling-based GNN training, distributed full-graph training allows learning over the complete input graphs, retaining whole graph structure information. Each device requires messages of all $1$-hop neighbors of nodes in its graph partition during iterative training, fetching the respective data from devices where they are stored/computed. The need for frequent message exchanges across devices renders the major performance bottleneck for training. Besides, different devices require different numbers of messages from others, which generates irregular all2all communications, leading to communication stragglers in each communication round. Such overheads of distributed full-graph training have also been echoed in recent literature~\cite{ wan2022pipegcn, peng2022sancus, Cai2021DGCLAE}. 

A few studies have investigated different perspectives to improve distributed full-graph training, including graph partition algorithms and memory management~\cite{Ma2019NeuGraphPD, Jia2020ImprovingTA}, communication planing~\cite{Cai2021DGCLAE} and communication-avoiding training with staleness~\cite{Thorpe2021DorylusAS, wan2022pipegcn, peng2022sancus}. Nevertheless, none of them considers compressing remote messages to reduce communication traffic for training expedition.\footnote{Model gradient compression has been extensively studied to accelerate distributed DNN training~\cite{Alistarh2017QSGDCS, Yu2019DoubleQF, Wu2018ErrorCQ}. However, for GNNs, the size of model gradients is typically much smaller than those of node features and embeddings (e.g., for the ogbn-products dataset, a three-layer GCN with a hidden size of 256 has 0.55MB model gradients, but 1.17GB features and 3.00GB embeddings), making the transferring of messages much more costly than that of gradients.} 
Unlike the above methods, message compression can reduce data volumes for both communications between remote devices and data movement from device to host. The latter occurs when messages need to be moved from device memory to host memory first and then transferred to remote devices. (when GPUDirect RDMA is not available in the cluster).

While messages are being exchanged, embedding computation of nodes with all neighbors located locally often waits for the completion of message transfers in synchronous full-graph training~\cite{Ma2019NeuGraphPD, Jia2020ImprovingTA, Cai2021DGCLAE}, which is not needed. Although existing staleness-based methods eliminate part of the waiting time by pipelining communication with computation~\cite{wan2022pipegcn} or skipping node broadcast and using historical embeddings for computation~\cite{peng2022sancus}, 
they may lead to slower training convergence~\cite{wan2022pipegcn, peng2022sancus}, increasing the wall-clock time to achieve the same model accuracy as compared to their synchronous counterparts. Disparate handling of local nodes with and without remote neighbors has not been found in both synchronous and asynchronous full-graph training works in the literature.


We propose AdaQP, an efficient distributed full-graph GNN training system that accelerates GNN training from two 
perspectives: adaptive quantization of messages and parallelization of computation of central nodes and message transfers of marginal nodes on each device. Our main contributions are summarized as follows:

$\triangleright$ We apply adaptive stochastic integer quantization to messages dispatched from each device to others, which reduces the numerical precision of messages and hence the size of transferred data. To our best knowledge, we are the first to apply stochastic integer quantization to expedite distributed full-graph training. We provide a theoretical convergence guarantee and show that the convergence rate is still $O(T^{-1})$, identical to that of no-compression training and better than sampling-based~\cite{cong2020minimal} and staleness-based~\cite{wan2022pipegcn, peng2022sancus} GNN training. Using insights from the convergence analysis, an adaptive bit-width assignment scheme is proposed based on a bi-objective optimization, that assigns suitable quantization bit-width to the transferred messages to alleviate 
unbalanced data volumes from/to different devices and achieve a good trade-off between training convergence and efficiency.

$\triangleright$ We further decompose the graph partition on each device into a \textit{central graph} and a \textit{marginal graph}, and overlap the computation time of the former with the message communication time of the latter, to maximize training speed and resource utilization. Since the communication overhead always dominates the training process (Sec.~\ref{sec:2overhead}), the computation time of the central graph can be easily hidden within the communication time without introducing any staleness that influences training convergence. %

$\triangleright$ We implement AdaQP on DGL~\cite{wang2019deep} and PyTorch~\cite{paszke2019pytorch}, and conduct extensive evaluation. Experimental results show that AdaQP significantly reduces the communication time by 80.94\% maximum and 79.98\% on average, improves the training throughput by 2.19 $\sim$ 3.01$\times$ with acceptable accuracy fluctuations (- 0.30\% $\sim$ + 0.19\%), and outperforms state-of-the-art (SOTA) works on distributed full-graph training on most of the mainstream graph datasets.

\section{Background and Motivation}\label{sec:B&M}


\subsection{GNN Message Passing}\label{sec:gnn_frame}

The message passing paradigm of GNN training can be described by two stages, aggregation and update~\cite{Gilmer2017NeuralMP}:

\vspace{-6mm}
\begin{equation}\label{eq:agg}
\small
    h_{N(v)}^{l} = \phi^{l}(h_u^{l-1} |u \in N(v))
\end{equation}
\vspace{-9mm}

\begin{equation}\label{req:update}
\small
    h_v^{l}=\psi^{l}(h_v^{l-1}, h_{N(v)}^l)
\end{equation}
\vspace{-7mm}

\noindent Here $N(v)$ denotes the neighbor set of node $v$. $h_v^{l}$ is the learned embedding of node $v$ at layer $l$. $\phi^{l}$ is the aggregation function of layer $l$, which aggregates intermediate node embeddings from $N(v)$ to derive the aggregated neighbor embedding $h_{N(v)}^{l}$. $\psi^{l}$ is the update function that combines $h_{N(v)}^l$ and $h_v^{l-1}$ to produce $h_v^{l}$. 


The two-stage embedding generation can be combined into a weighted summation form, representing the cases in most mainstream GNNs (e.g., GCN~\cite{Kipf2017SemiSupervisedCW}, GraphSAGE~\cite{hamilton2017inductive}):

\vspace{-5mm}
\begin{equation}\label{eq:gnn_sum}
\small
    h_{v}^l = \sigma(W^l\cdot (\sum_{u}^{\{v\} \cup N(v)} \alpha_{u,v}h_u^{l-1}))
\end{equation}
\vspace{-5mm}


\noindent where $\alpha_{u,v}$ is the coefficient of embedding $h_u^{l-1}$, 
$W^l$ is the weight matrix of layer $l$ and $\sigma$ is the activation function. We will use this form in our theoretical analysis in Sec.~\ref{sec:theory}.

\subsection{Inefficient Vanilla Distributed Full-graph Training}\label{sec:2overhead}

\begin{figure}[!th]
  \centering
  \includegraphics[width=\linewidth]{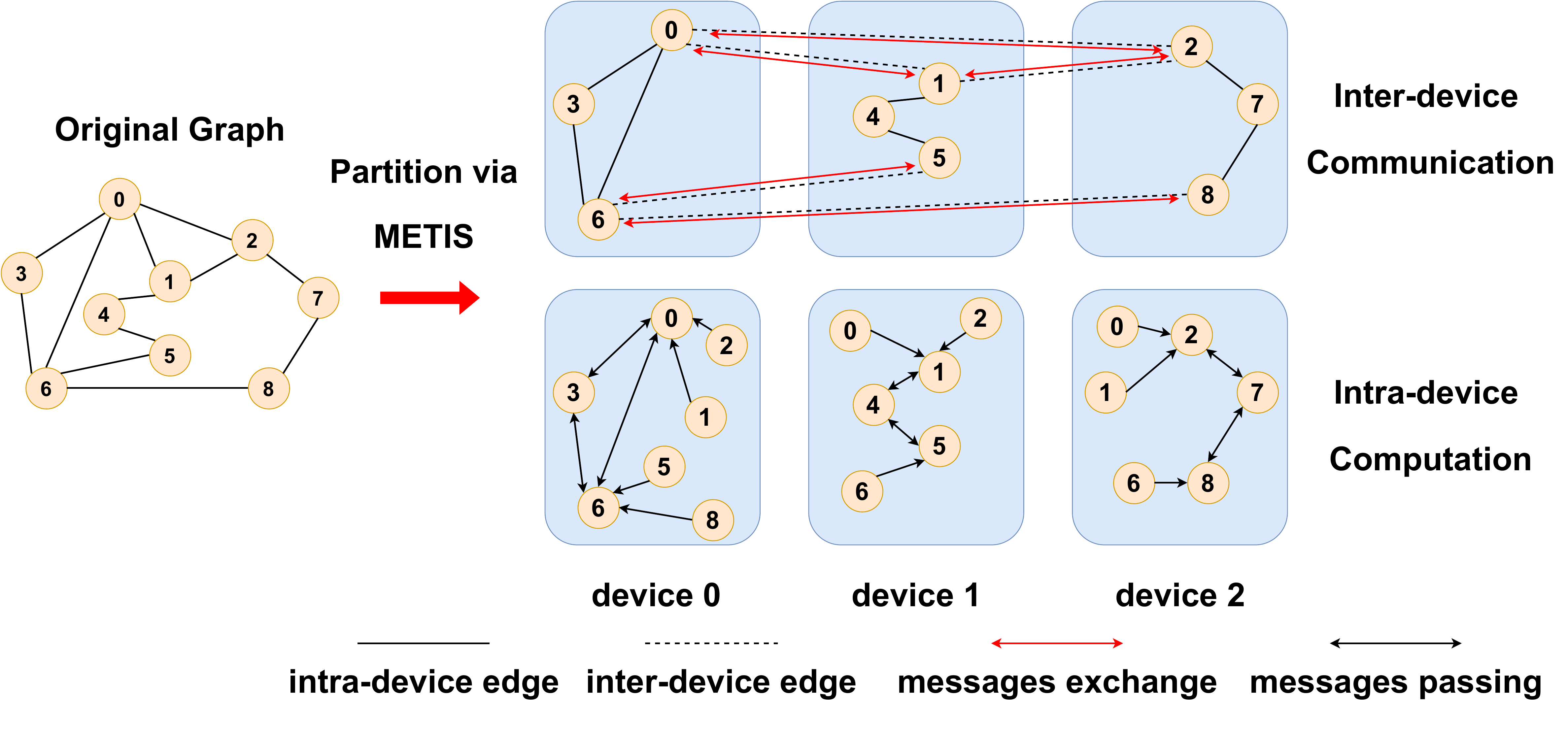}
  \vskip -0.15in
  \caption{An illustration of vanilla distributed full-graph training. The original graph is partitioned into 3 partitions by METIS~\cite{karypis1997metis} 
  and deployed on 3 devices. Inter-device communication and intra-device computation are repeated in the forward (backward) pass of each GNN layer.
  }
  \label{fig:illustra}
  \vspace{-5mm}
\end{figure}


\begin{table}[!th]
\caption{Communication overhead in Vanilla. $x$M-$y$D means the whole graph is partitioned into $x\times y$ parts and dispatched to $x$ servers, each using $y$ available training devices (GPUs). 
  }
\label{tab:com}
\vskip 0.1in
\resizebox{\columnwidth}{!}{
\begin{tabular}{cccc}
    \toprule
    Dataset & Partition Setting & Communication Cost & Remote Neighbor Ratio \\
    \midrule
    \multirow{2}*{Reddit~\cite{hamilton2017inductive}}& 2M-1D & 66.78\% & 41.54\% \\
     & 2M-2D & 75.20\%   & 62.60\% \\
     \hline
     \multirow{2}*{ogbn-products~\cite{hu2020open}}& 2M-2D & 75.59\% & 31.09\% \\
     & 2M-4D & 76.67\% & 40.52\% \\
     \hline
     \multirow{2}*{AmazonProducts~\cite{Zeng2020GraphSAINTGS}} & 2M-2D & 75.58\% & 39.75\% \\
     & 2M-4D & 78.22\% & 53.00\%  \\
    \bottomrule
\end{tabular}}
\end{table}

In vanilla distributed full-graph training (Fig.~\ref{fig:illustra}, referred to as \textit{Vanilla}), interleaving communication-computation stages exist in each GNN layer during both forward pass and backward pass for generating embeddings or embedding gradients. Therefore, multiple transfers of 1-hop remote neighbors' messages (specifically, transferring features and embeddings in the forward pass and embedding gradients, also denoted as 
\textit{errors}~\cite{goodfellow2016deep}, in the backward pass) lead to large communication overhead. To illustrate it, we train a three-layer GCN on representative datasets (all experiments in this section use this GCN, detailed in Sec.~\ref{sec:exp}) and show the communication cost, which is computed by dividing the average 
communication time by the average per-epoch training time among all devices, in Table~\ref{tab:com}. We observe that communication time dominates training time. Further, with the increase of partition number, the communication cost becomes larger due to the growth of the remote neighbor ratio (computed by dividing the average number of remote $1$-hop neighbors by the average number of nodes among partitions).

\begin{figure}[!t]
\vspace{-3mm}
  \centering
  \includegraphics[height=0.6\linewidth]{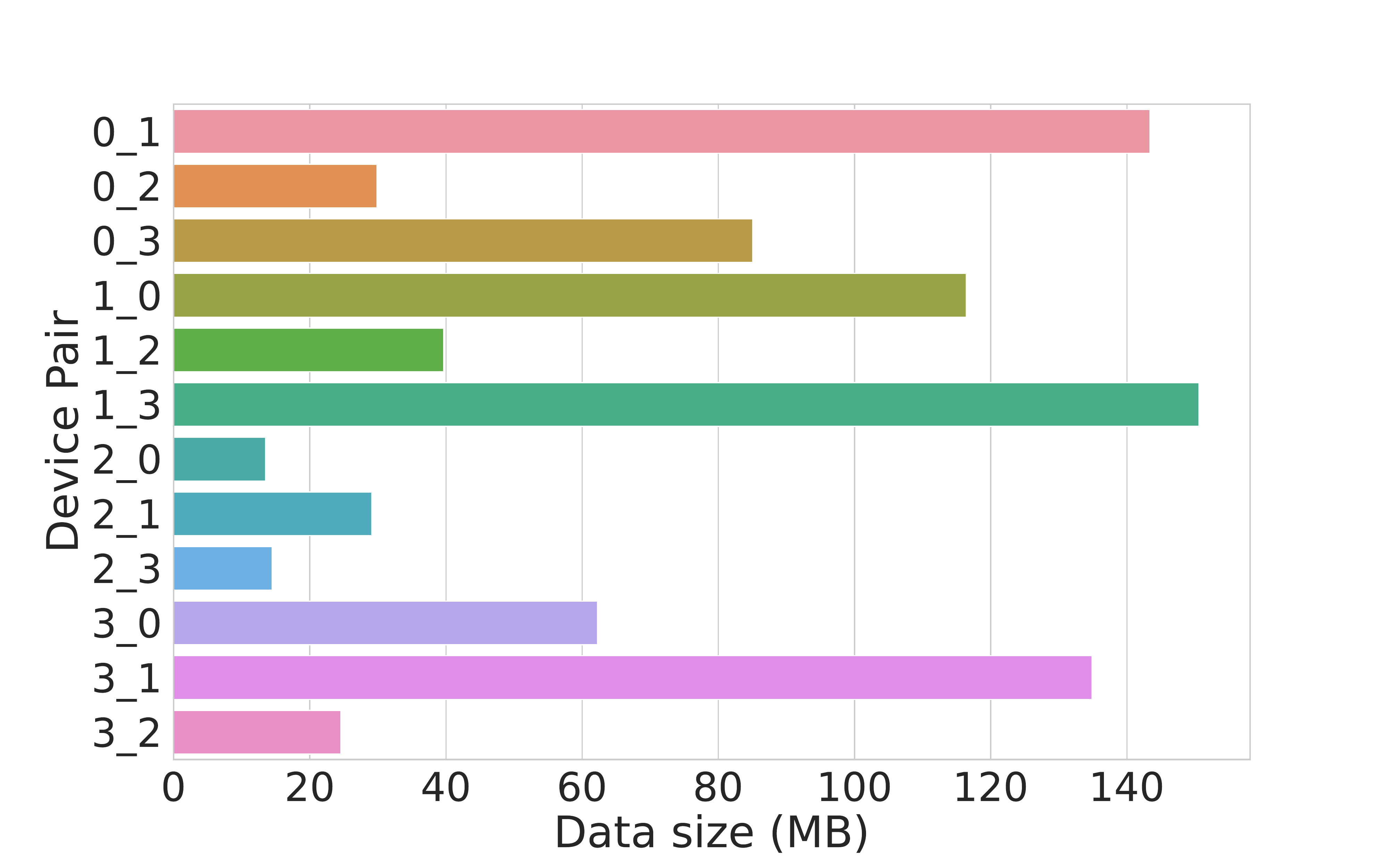}
  \vspace{-3mm}
  \caption{Comparison of data size transferred across each device pair 
  when training the GCN on AmazonProducts with 4 partitions. 
  }
  \label{fig:irregular}
  \vspace{-7mm}
\end{figure}

Besides, 
with the mainstream graph partition algorithms (e.g., METIS~\cite{karypis1997metis}
), the number of nodes whose messages are transferred 
varies among different device pairs, creating 
unbalanced all2all communications. This unique communication pattern exists throughout the GNN training process, which does not occur in distributed DNN training (where devices exchange same-size model gradients). Fig.~\ref{fig:irregular} shows the data size transferred across different device pairs in GCN's first layer 
when training on AmazonProducts, partitioned among 4 devices. There is a significant imbalance among data sizes transferred across different device pairs, which leads to unbalanced communication time across the devices 
in each communication round, affecting the overall training speed.

\begin{figure}[!t]
    \centering
    \includegraphics[width=\linewidth]{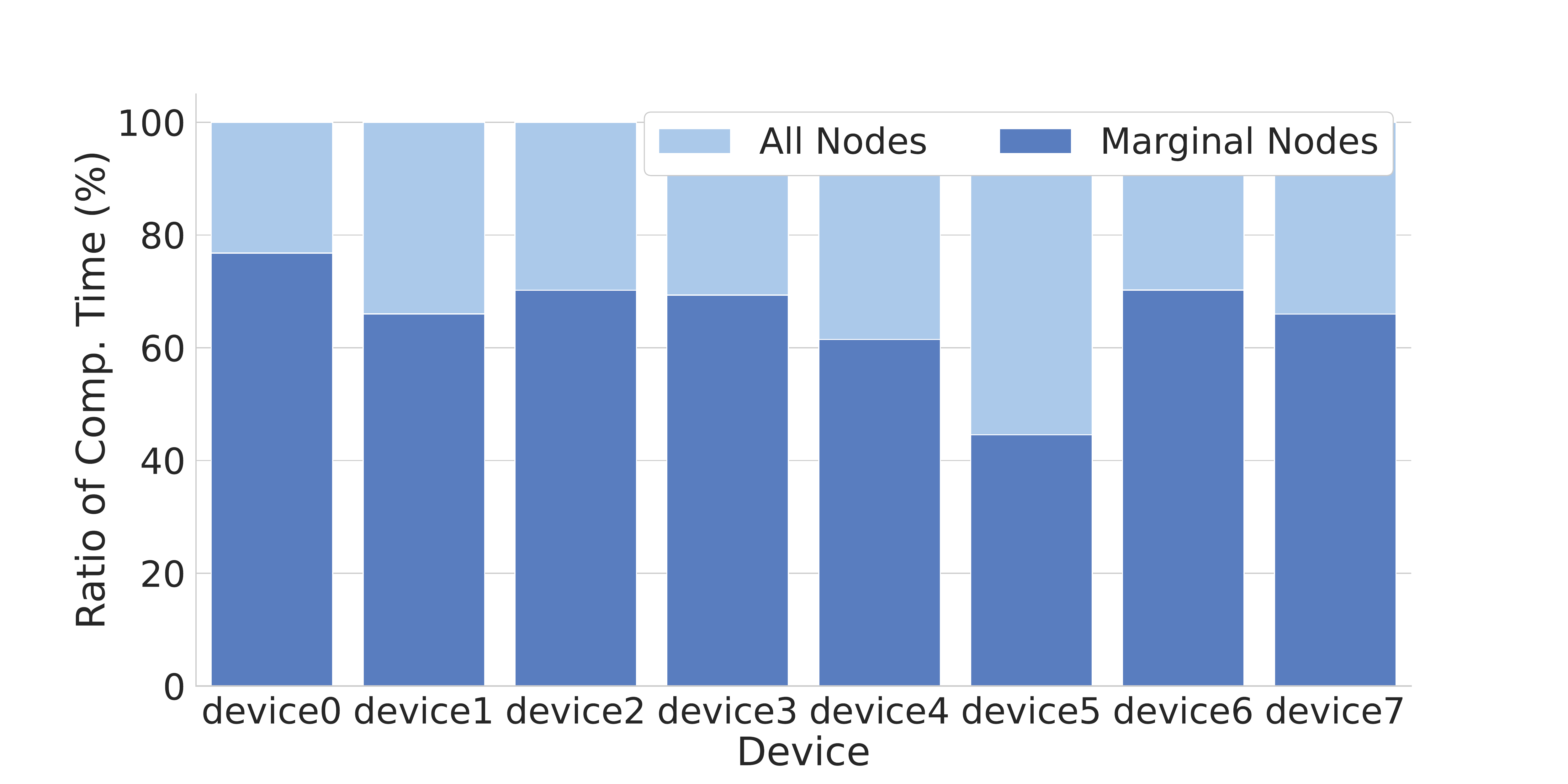}
    \vspace{-7mm}
    \caption{Comparison between computation time of all nodes and computation time of marginal nodes 
    when training on ogbn-products with 8 partitions. 
    } 
    \label{fig:comp_ratio}
\end{figure}

Further, Vanilla and previous works~\cite{wan2022pipegcn, peng2022sancus, Cai2021DGCLAE}
do not consider that in the forward (backward) pass of each training iteration, the computation of central nodes can  directly start without waiting for message exchanges. Overlapping the computation time of central nodes with the communication time of marginal nodes can help further improve the training throughput. 
As communication renders the major bottleneck in GNN training, we observed that central nodes' computation time can be well hidden within marginal nodes' communication time. In Table~\ref{tab:comm_compu_compare}, we show central nodes' computation time and marginal nodes' communication time when the transferred messages are quantized with a bit-width of $2$ (i.e., 
the numerical precision is 2-bit), rendering the lowest communication volumes. 
Even under this extremely low communication, communication time is still longer than the central nodes' computation time. 
When central node computation is hidden within communication, Fig.~\ref{fig:comp_ratio} shows the reduction of model computation time on each device, by 23.20\% to 55.44\%. 

\begin{table}[!t]
\caption{Computation time of central nodes and transfer time of 2-bit quantized messages of marginal nodes on ogbn-products with 8 partitions.}
\label{tab:comm_compu_compare}
\vskip 0.1in
\begin{small}
\resizebox{\linewidth}{!}{
\begin{tabular}{ccc|ccc}
    \toprule
    \textbf{Device} & \textbf{comm. (s)} & \textbf{Comp. (s)} & \textbf{Device} & \textbf{comm. (s)} & \textbf{Comp. (s)} \\
    \midrule
     Device0 & 0.08 & 0.04 & Device4 & 0.08 & 0.06 \\
     Device1 & 0.09 & 0.05 & Device5 & 0.13 & 0.06 \\
     Device2 & 0.10 & 0.05 & Device6 & 0.09 & 0.06 \\
     Device3 & 0.08 & 0.05 & Device7 & 0.12 & 0.05\\
    \bottomrule
\end{tabular}}
\end{small}
\vspace{-5mm}
\end{table}

\subsection{Stochastic Integer Quantization}
As a lossy compression method, stochastic integer quantization~\cite{chen2021actnn} has been applied to quantize the DNN model for efficient integer arithmetic inference~\cite{zhu2020towards, Tailor2021DegreeQuantQT, Feng2020SGQuantST}, or compress activations to reduce memory footprint during the forward pass~\cite{chen2021actnn, liu2021exact}. Differently, we apply it to reduce communication data volumes in distributed full-graph training. For a given message vector $h_{v}^l$ of node $v$ in layer $l$, the $b_v$-bit quantized version of $h_{v}^l$ is:
\vspace{-4mm}
\begin{equation}\label{eq:quant}
\small
    \tilde{h}_{v_b}^l = \tilde{q_b}(h_v^l)=round_{st}(\frac{h_v^l-Z_v^l}{S_{v_b}^l})
\vspace{-4mm}
\end{equation}

where $\tilde{q_b}$ denotes stochastic integer quantization operation, $round_{st}(\cdot)$ is the stochastic rounding operation~\cite{chen2020statistical}, $Z_v^l=\min(h_v^l)$ is referred to as the zero-point (the minimum value among elements in vector $h_{v}^l$) and $S_{v_b}^l=\frac{max(h_v^l)-min(h_v^l)}{2^{b_v}-1}$ is a scaling factor that maps the original floating-point vector into the integer domain,
where $b_v$ is typically chosen among $\{2,4,8\}$. A larger quantization bit-width $b_v$ introduces less numerical precision loss, but not as much data size reduction as a smaller value. Received quantized messages are de-quantized into floating-point values for subsequent computation:
\vspace{-2mm}
\begin{equation}\label{eq:dequant}
\small
    \hat{h}_v^l=dq_b(\tilde{h}_{v_b}^l)=\tilde{h}_{v_b}^l S_{v_b}^l + Z_v^l
\vspace{-5mm}
\end{equation}

\noindent where $\hat{h}_v^l$ represents the de-quantized message vector and $dq_b$ denotes de-quantization operation. 
Following~\cite{chen2021actnn}, $\hat{h}_{v}^{l}$ 
is \textit{unbiased} and \textit{variance bounded}:
\begin{theorem}\label{thm:quant}
\small
With stochastic integer quantization and deterministic de-quantization operations $\tilde{q}_b(\cdot)$ and $dq_b(\cdot)$ in Eqn.~(\ref{eq:quant}) and Eqn.~(\ref{eq:dequant}), $\hat{h}_{v}^{l}$ is an unbiased and variance bounded estimate of the original input $h_{v}^{l}$, that $\mathbb{E}[\hat{h}_{v}^{l}]=\mathbb{E}[dq_b(\tilde{h}_{v_b}^{l})]=h_{v}^{l},\mathbb{V}ar[\hat{h}_{v}^{l}]=\frac{D_v^{l}(S_{v_b}^{l})^2}{6}$. $D_v^{l}$ is the dimension of vector $h_v^{l}$.
\end{theorem}

The good mathematical properties of the quantization method serve as the basis for us to derive a theoretical guarantee of GNN training convergence (Sec.~\ref{sec:theory}).

\begin{figure}[!t]
    \centering

    \subfigure[Vanilla]{
    \includegraphics[width=\linewidth]{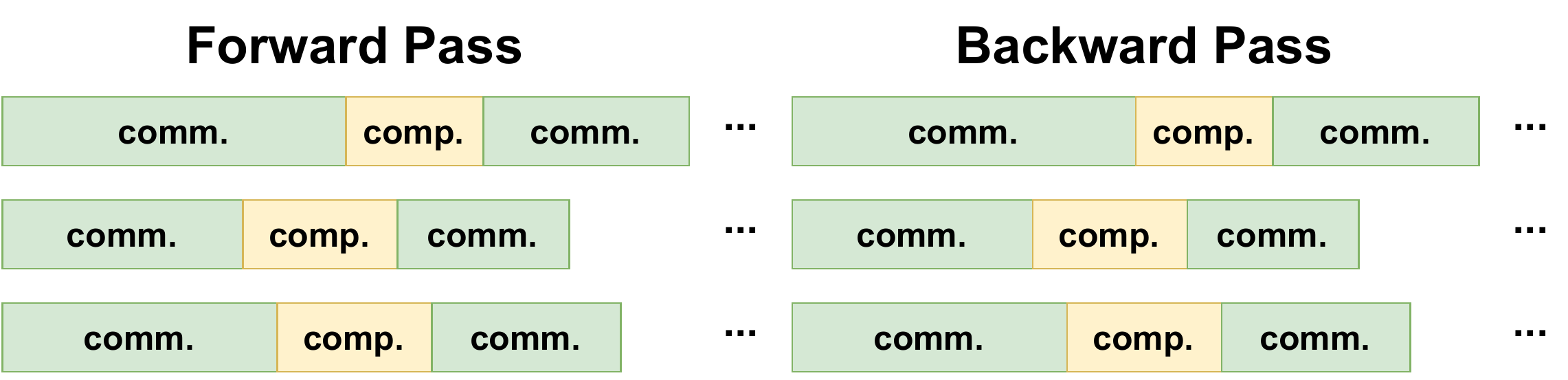}
    \label{fig:vanilla}
    }
    \subfigure[AdaQP]{
    \includegraphics[width=\linewidth]{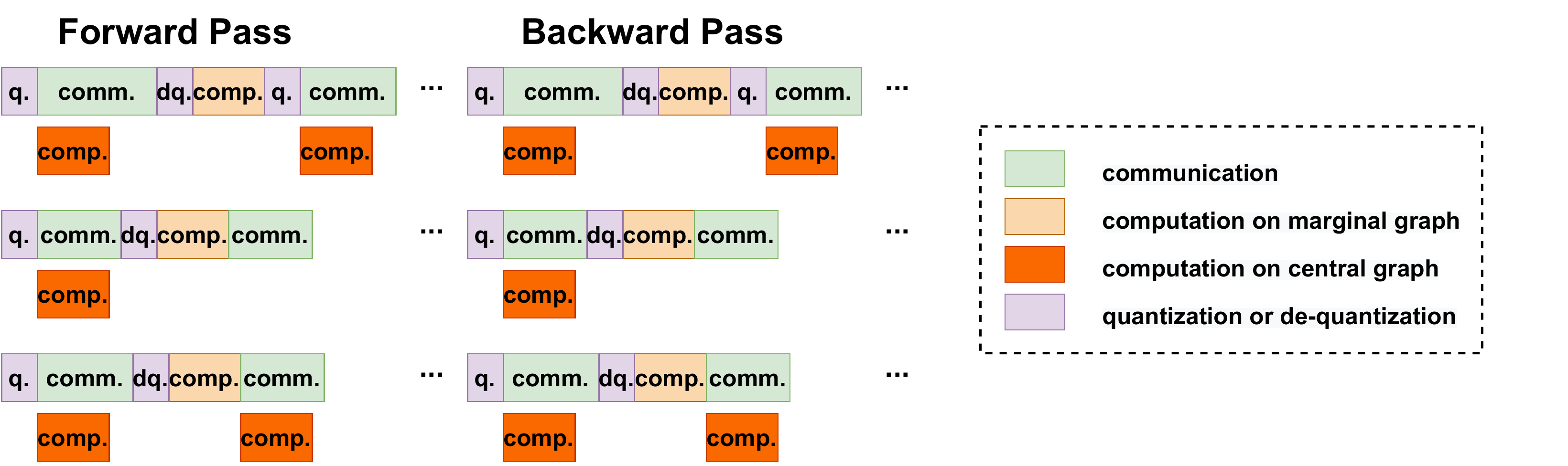}
    \label{fig:ours}
    }
    \vspace{-5mm}
    \caption{Comparison of Vanilla and AdaQP. 
    }
    \label{fig:high_compare}
\end{figure}

We propose an efficient system AdaQP, incorporating adaptive message quantization and computation-communication parallelization design, to improve distributed full-graph training efficiency. Fig.~\ref{fig:high_compare}  gives a high-level illustration of the benefits of AdaQP, as compared to Vanilla. 

\section{System Design}\label{sec:design}

\begin{figure}[!t]
  \centering
  \includegraphics[width=\linewidth]{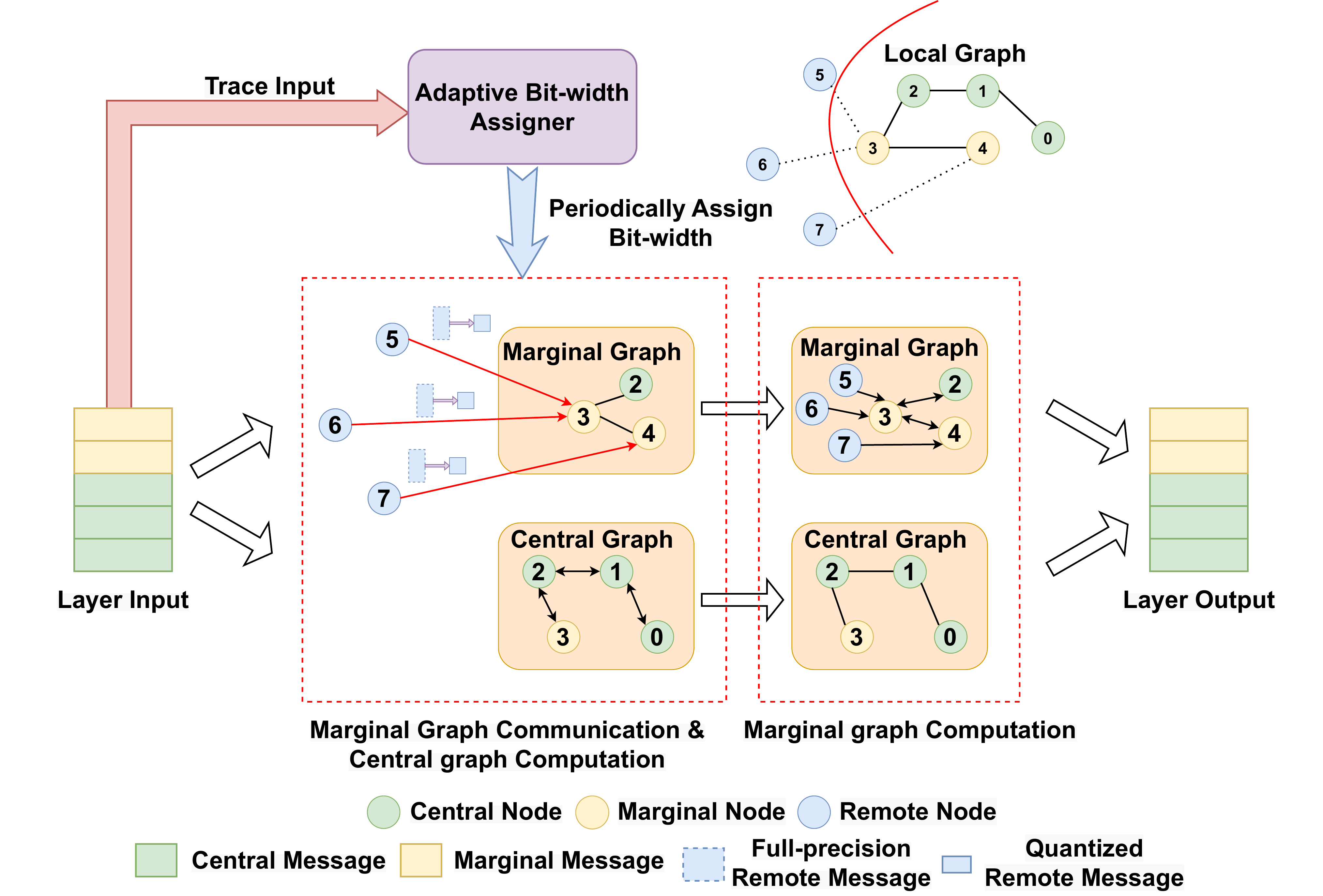}
  \vspace{-7mm}
  \caption{Workflow of AdaQP on each device: a per-layer view of GNN training. Only the messages receiving process is illustrated for clarity of the figure. 
  }
  \label{fig:workflow}
  \vspace{-3mm}
\end{figure}

We study distributed full-graph GNN training using devices (aka workers) on multiple physical machines. The large input graph is partitioned among the devices. Fig.~\ref{fig:workflow} shows the workflow of distributed full-graph training on each device using AdaQP. 

\subsection{Overview}

The graph partition at each device is decomposed into a \textit{central graph}, which contains central nodes and their neighbors, and a \textit{marginal graph}, which includes marginal nodes and their local neighbors. Messages from other devices only need to be passed to the marginal graph for computation. In the forward pass (backward pass) of each GNN layer, for nodes in the marginal graph, quantization is applied to all outgoing messages to other devices, 
using a bit-width assigned by the Adaptive Bit-Width Assigner, and de-quantization is done on all received messages before the computation stage begins. For central nodes in the central graph, they can enter their computation stage in parallel with the communication for the marginal graph. The Adaptive Bit-width Assigner traces the input data to each GNN layer, and periodically adjusts the bit-width for each message by solving a bi-objective problem with the latest traced data.


\subsection{Naive Message Quantization}\label{sec:quant_basic}
We perform stochastic integer quantization on the messages transferred across devices during the forward (backward) pass. Consider the aggregation step of GNN layer $l$ for computing embedding of a node $v$, as given in Eqn.~\ref{eq:agg}
in Sec.~\ref{sec:gnn_frame}. Let $N_L(v)$ and $N_R(v)$ denote the local and remote neighbor sets of node $v$. We rewrite the aggregation step as:

\vspace{-4mm}
\begin{equation}
\small
    h_{N(v)}^{l} = \phi^{l}(h_u^{l-1} |u \in \{N_L(v) \cup N_R(v)\})
\end{equation}
\vspace{-7mm}

Quantization is only performed on messages in $N_R(v)$
to reduce the communication data size. 
The benefit brought by naive message quantization is obvious; however, it requires frequently performing quantization and de-quantization operations before and after each device-to-device communication, which adds extra overheads to the training workflow. We notice that quantization (Eqn.~\ref{eq:quant}) and de-quantization (Eqn.~\ref{eq:dequant}) themselves are particularly suitable for parallel computing since they just perform the same linear mapping operations to all elements in a message vector. Therefore, we implement the two operations with efficient CUDA kernels. Sec~\ref{sec:exp_break} demonstrates that compared to training expedition brought by quantization, the extra overheads are almost negligible.  

\subsection{Dynamic Adaptive Bit-width Assignment}\label{sec:quant_adaptive}

\begin{figure}[!t]
  \centering
  \includegraphics[width=\linewidth]{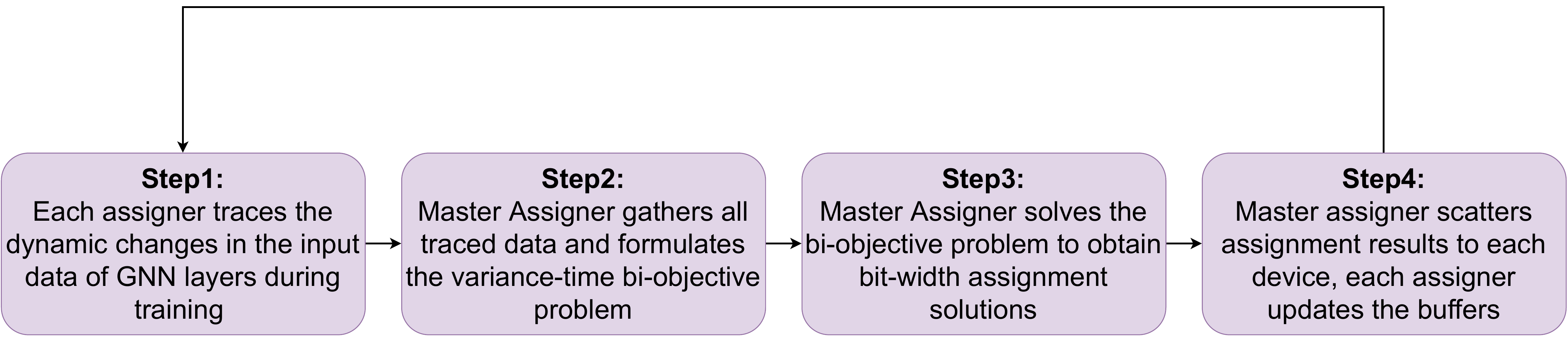}
  \caption{Adaptive bit-width assignment process. 
  }
  \label{fig:adaptive_assign}
\end{figure}

Simply assigning the same bit-width to all messages cannot achieve a good tradeoff between training convergence and training efficiency (Sec.~\ref{sec:theory}); adaptively assigning different bit-widths to different messages is necessary. To support transferring messages quantized with multiple bit-widths between devices, each device first establishes multiple sending buffers for messages of different bit-widths, whose sizes are determined by the Adaptive Bit-width Assigner, and then broadcasts the sizes of the buffers to other devices. Each device also uses received sending buffer sizes from others to set up multiple 
receiving buffers. The adaptive bit-width assigning process is shown in Fig.~\ref{fig:adaptive_assign}. Given a bit-width update period, each training device launches an Adaptive Bit-width Assigner which keeps tracing all GNN layers' inputs. The assigner periodically sends traced data from the last period to the master assigner (residing in the device with rank 0) and then blocks the current training worker, waiting for the generation of new bit-width assignment results (step2). At the same time, the master assigner uses gathered data to build a variance-time bi-objective problem, whose variables are bit-widths of the message groups. 
Since the bit-width assignment results for each GNN layer have no dependence on each other, we create a thread pool in the master device to compute each layer's solution in parallel (step 3). After that, the master assigner scatters the bit-width solutions to corresponding devices, and then each device uses the latest assignment results to update its sending and receiving buffers (step 4). 

\subsection{Computation-Communication Parallelization}\label{sec:para}

Model computation and message quantization are both compute-intensive. Therefore overlapping computation on the central graph with the quantized message transfers on the marginal graph can lead to an overall slow-down due to the contention for GPU compute resources, which is indicated in previous works~\cite{agarwal2022utility}. To tackle this issue, we apply a simple yet useful resource isolation strategy based on the observations in Sec.~\ref{sec:2overhead}, that is, the time for transferring extremely quantized messages is still large enough to hide the computation time of the central graph. We control CUDA kernel launching time instead of letting GPUs schedule different CUDA streams freely.

\begin{figure}[!t]
  \centering
  \includegraphics[height=0.6\linewidth]{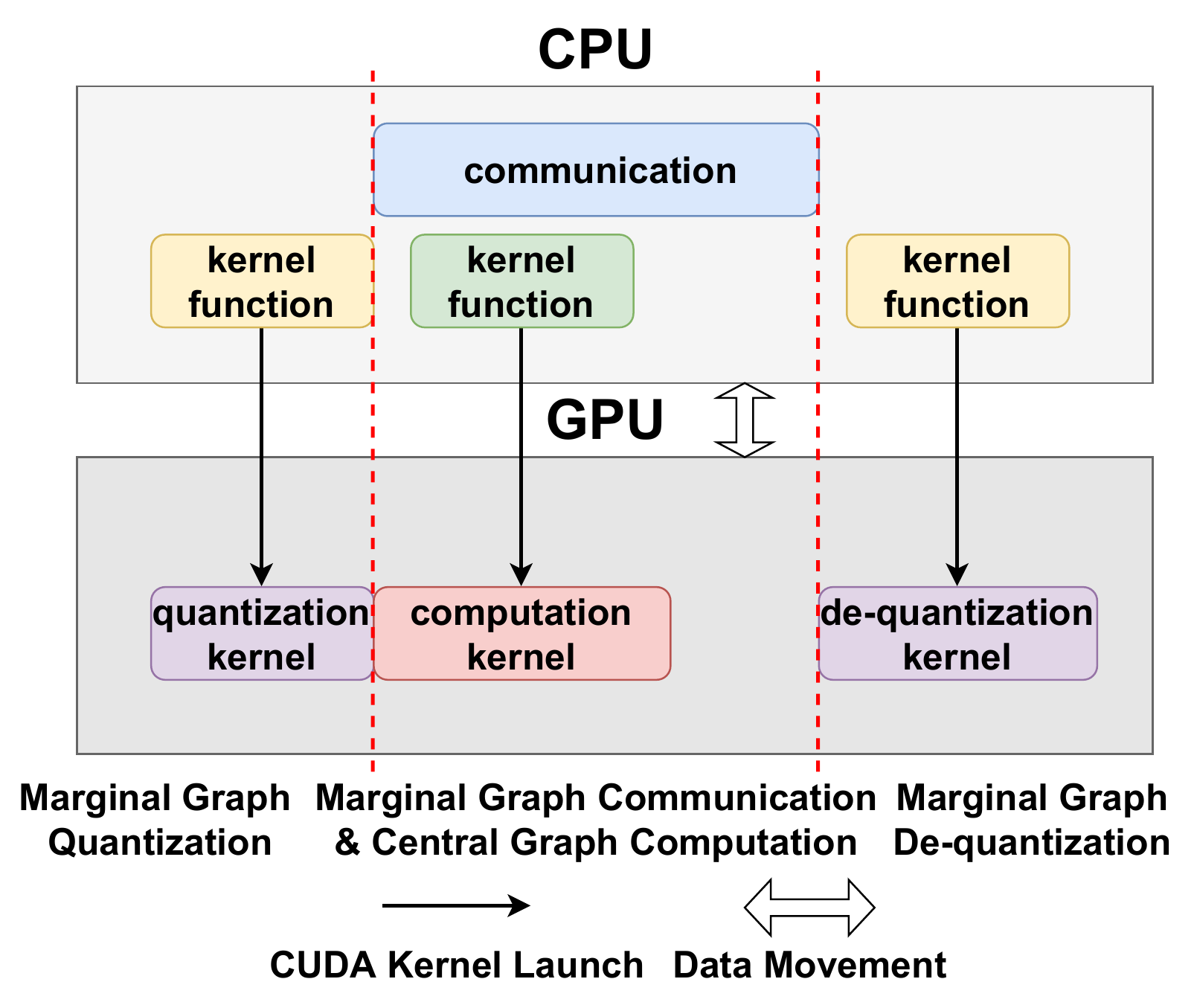}
  \vspace{-3mm}
  \caption{Our GPU resources isolation strategy. 
  }
  \label{fig:isolation}
  \vspace{-3mm}
\end{figure}

As shown in Fig.~\ref{fig:isolation}, we further divide the computation-communication overlapping stage in Fig.~\ref{fig:workflow} into three fine-grained stages, ensuring that in each stage the GPU compute resources are only used by one among quantization, de-quantization and central graph computation. 
Since communication on the marginal graph only requires the GPU bandwidth, we force the computation CUDA kernel launching and execution on the central graph to be in the second stage with marginal-graph communication, isolating the GPU resource usage between central and marginal graphs. We wrap different types of kernels in GPU with different CUDA streams and consider the situation that CUDA kernel's launching and execution are asynchronous. We use both CPU and GPU events for synchronization.

\section{Convergence Guarantee and Bi-objective Bit-width Assignment}\label{sec:theory}

We next show the convergence bound and rate of AdaQP and then formulate a variance-time bi-objective optimization problem for adaptive bit-width assignment.

\subsection{Impact of Message Quantization on Training Convergence}\label{sec:conver}

We consider widely adopted gradient descent (GD) algorithm~\cite{avriel2003nonlinear} in full-graph GNN training.
Similar to previous studies~\cite{fu2020don, chen2020statistical}, the full-graph training can be modeled as the following non-convex empirical risk minimization problem:

\vspace{-8mm}
\begin{equation}
\small
    \min_{\mathbf{w}_t\in R^D} \mathbb{E}[\mathcal{L}(\mathbf{w}_t)] = \frac{1}{N}\sum_i^N \mathcal{L}_i(\mathbf{w}_t)\ \ \ \mathbf{w}_{t+1} = \mathbf{w}_t - \alpha \mathbf{\tilde{g}}_t
\end{equation}
\vspace{-7mm}

\noindent where $\mathbf{w}_t$ denotes the parameters at the $t$-th epoch, $D$ is the dimension of $\mathbf{w}_t$, $N$ denotes the total number of samples (nodes) in the full-graph, and $\mathcal{L}_i(\cdot)$ is the loss on sample $i$. 
$\mathbf{\tilde{g}}_t=\nabla \mathcal{L}(\mathbf{\tilde{w}}_t)$ denotes the stochastic gradient and $\alpha$ is the learning step size. Since each training epoch uses full-batch samples, variance in $\mathbf{\tilde{g}}_t$ is only brought by performing stochastic integer quantization on messages.
We have Theorem~\ref{thm:conver} based on standard assumptions for convergence analysis~\cite{fu2020don, allen2017natasha}.

\begin{assumption}\label{assum:conver}
\small
for $\forall \mathbf{w}_t, \mathbf{w'}_t \in R^D$ in the $t$-th epoch:

\textbf{A.1} ($\mathcal L_2-Lipschitz$) $||\nabla \mathcal{L}(\mathbf{w}_t)-\nabla \mathcal{L}(\mathbf{w'}_t)|| \leq L_2||\mathbf{w}_t-\mathbf{w'}_t||$;

\textbf{A.2} (\emph{existence of global minimum}) $\exists\ \mathcal{L}^{\ast}\ s.t.\ \mathcal{L}(\mathbf{w}_t) \ge \mathcal{L}^{\ast}$;

\textbf{A.3} (\emph{unbiased stochastic gradient}) $\mathbb{E}[\mathbf{\tilde{g}}_t]= \mathbf{g}_t;$

\textbf{A.4} (\emph{bounded variance}) $\mathbb{E}[||\mathbf{g}_t-\tilde{\mathbf{g}}_t||] \leq Q.$
\end{assumption}
\begin{theorem}\label{thm:conver}
\small
Suppose our distributed full-graph GNN training runs for $T$ epochs using a fixed step size $\alpha < \frac{2}{L_2}$. Select $t$ randomly from $\{1, \cdots, T\}$. Under Assumption~\ref{assum:conver}, we have

\vspace{-3mm}
\begin{equation}
    \mathbb{E}[||\nabla \mathcal{L}(\bar{\mathbf{w}}_t)||^2] \leq 
\frac{2(\mathcal{L}(\mathbf{w}_1)-\mathcal{L}^{\ast})}{T(2\alpha - \alpha^2L_2)} + \frac{\alpha L_2Q^2}{2-\alpha L_2}.
\end{equation}
\vspace{-5mm}

\end{theorem}

All the detailed proofs can be found in Appendix~\ref{app:proof}. Similar to the convergence result of a standard SGD algorithm~\cite{robbins1951stochastic}, the bound in Theorem~\ref{thm:conver} includes two terms, while  the second term is totally different from its SGD counterpart, where variance is not introduced due to sampling variance but message quantization. The first term in the bound goes to $0$ as $T\rightarrow \infty$, which shows an $O(T^{-1})$ convergence rate. AdaQP's convergence rate is the same as that with Vanilla, and faster than those of sampling-based methods ($O(T^{-\frac{1}{2}})$~\cite{cong2020minimal}) and staleness-based methods ($O(T^{-\frac{2}{3}})$~\cite{wan2022pipegcn} or $O(T^{-\frac{1}{2}})$~\cite{peng2022sancus}). 

During training, the gradient variance exists in the weight matrix in each layer $l$ of an $L$-layer GNN. Let $\mathbf{w}=\{\mathbf{w}^l\}^{L}_{l=1}$ denotes the set of weight matrices of GNN,
we then show the gradient variance upper bound $Q^l$ for each $\mathbf{w}^l$. Let $\bar{h}_v^{l-1}=\sum_{u}^{\{v\} \cup N(v)} \alpha_{u,v}h_u^{l-1}$ in Eqn.~\ref{eq:gnn_sum}, and $\frac{\partial \bar{\mathcal{L}}}{\partial h_v^l}=\sum_{u}^{\{v\} \cup N(v)} \alpha_{u,v}\frac{\partial \mathcal{L}}{\partial h_u^l}$ in its backward pass counterpart. 
Based on Theorem~\ref{thm:quant}, we derive $Q^l$ under Assumption~\ref{assum:layer}:

\begin{assumption}\label{assum:layer}
\small
For each layer $l \in \{1,2,\cdots,L\}$ in the GNN and for each node $v$ in the full-graph,  $L_2$ norms of the expectations of $\bar{h}_v^{l-1}$ and $\frac{\partial \bar{\mathcal{L}}}{\partial h_v^l}$ are upper-bounded: $||\mathbb{E}[\bar{h}_v^{l-1}]|| \leq M,\ ||\mathbb{E}[\frac{\partial \bar{\mathcal{L}}}{\partial h_v^l}]|| \leq N$. 
\end{assumption}

\begin{theorem}\label{thm:layer}
\small
Given a distributed full-graph $(V, E)$ and optional bit-width set $
B$, for each layer $l \in \{1, \cdots, L\}$ in the GNN, gradient variance upper bound $Q^l$ in layer $l$ is:
\begin{equation}
\small
    \begin{aligned}
          & Q^l = \sum_v^{|V|}(\sum_{k_1}^{N_R(v)}\sum_{k_2}^{N_R(v)}\alpha^2_{k_1,v}\alpha^2_{k_2,v} \frac{D_{k_1}^{l-1}D_{k_2}^l(S_{k_{1_b}}^{l-1}S_{k_{2_b}}^l)^2}{6}\\
         &+ M^2 \sum_k^{N_R(v)}\alpha^2_{k,v} \frac{D_k^{l}(S_{k_b}^{l})^2}{6}
         + N^2 \sum_k^{N_R(v)}\alpha^2_{k,v} \frac{D_k^{l-1}(S_{k_b}^{l-1})^2}{6})
    \end{aligned}
\end{equation}
\end{theorem}

\vspace{-2mm}
where the definitions of $S_{k_b}^{l}$ and $D_k^l$ can be found in Sec.~\ref{sec:quant_basic}. From a high-level view, for any $v$ in the distributed full-graph $(V, E)$, its neighborhood aggregation adds variance to model gradients in each layer if it has remote neighbors. For layer $l$, $Q^l$ is decided by many factors: (i) \textbf{graph topology and partition strategy}, which determine the size of $v$'s remote neighborhood $N_R(v)$; (ii) {\bf GNN aggregation function}, corresponding to the aggregation coefficient $\alpha_{k,v}$ for each node; (iii) \textbf{dimension size and numerical range of remote message vectors} ($D_k^{l}$, the numerator in $S_{k_b}^{l}$); (iv) \textbf{choices of quantization bit-width $b_v$} for each node. Given factors (i)-(iii) which are decided by the GNN training job, we can choose (iv) the quantization bit-width accordingly to minimize the terms in the bound, and thus reduce the gradient variance and let training converge closer to the solution of Vanilla~\cite{chen2021actnn}.
\subsection{Bi-objective Optimization for Adaptive Bit-width Assignment}
\label{sec:algorithm}

There is a trade-off in quantization bit-width assignment: 
using a larger quantization bit-width (e.g., 8-bit) leads to lower gradient variance upper bound $Q^l$ in each layer, but less communication volume reduction (as compared to using 4-bit or 2-bit). Our goal is to design an adaptive bit assignment scheme for different messages between each device pair, to strike a good balance between training convergence and efficiency. 

\begin{figure}[!t]
    \centering
    \includegraphics[width=\linewidth]{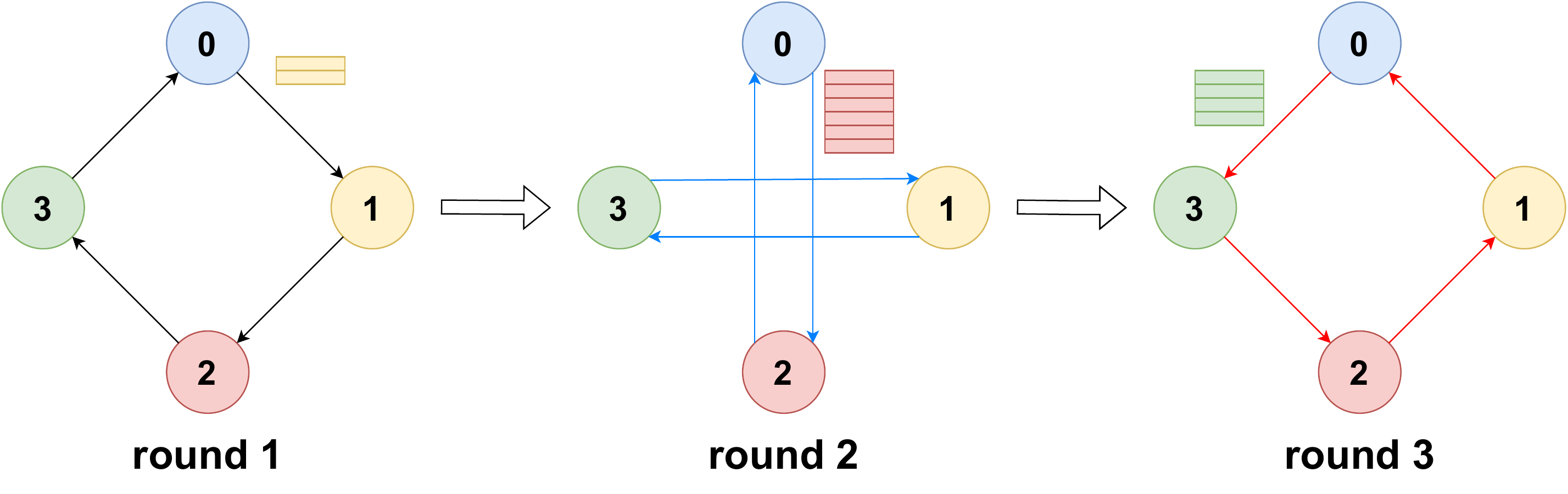}
    \vspace{-4mm}
    \caption{Illustration of ring all2all communication between four training devices. We only depict the data-sending process of device 0 for figure clarity.} 
    \label{fig:comp_ring}
    \vspace{-5mm}
\end{figure}

From Fig.~\ref{fig:irregular} we know that the size of transferred data varies significantly across device pairs. We should also 
alleviate communication stragglers in each communication round. Specifically, we follow ~\cite{wan2022pipegcn} to implement the all2all communication in a ring pattern (Fig.~\ref{fig:comp_ring}).
For $N$ devices, it takes $N-1$ communication rounds to finish the message exchange, where each of the devices sends/receives messages to/from its $i$-hop right/left neighbors in the ring during $i$-round communication.
Let $B = \{2, 4, 8\}$ be the set of candidate bit-widths. 
For each communication round in the forward (backward) pass of GNN layer $l$, we formulate the following minimax optimization problem for bit-width selection:

\vspace{-8mm}
\begin{equation}\label{eq:time}
\small
    \min_{b_k \in B} \max_{1\leq i \leq N} \theta_i \sum_{k}^{K_i} D_k^l b_k + \gamma_i
\end{equation}
\vspace{-4mm}

$b_k$ denotes the bit-width assigned for quantizing message $h_k^l$. For any device pair $i$, $K_i$ denotes the total number of messages to be transferred. $D_l^k$ is the dimension of the remote message vector $h_k^l$. $\theta_i$ and $\gamma_i$ are parameters of the cost model~\cite{sarvotham2001connection}. Problem~\ref{eq:time} finds the device pair that has the longest predicted communication time and minimizes it. 

According to Theorem~\ref{thm:layer}, considering a message $h_k^l$ in GNN layer $l$ sent to a target device, we want to minimize the variance it brings to this layer's weight gradients. Apart from bit-width $b_k$, the dimension and the minimum (maximum) values of it (in $S_{k_b}^l$), 
and the sum of squares of all the aggregation coefficients $\sum_v^{N_{T}(k)}\alpha^2_{k,v}$ (where $N_{T}(k)$ denotes $k$'s neighbors in the target device) allocated by its neighbors in the target device also influence the magnitude of variance. We solve the following minimization problem 
to minimize the total gradient variance in one communication round, where $\beta_k=\frac{\sum_v^{N_{T}(k)}\alpha^2_{k,v}D_k^l(max(h_k^l)-min(h_k^l))^2}{6}$:

\vspace{-4mm}
\begin{equation}\label{eq:variance}
\small
    \min_{b_k \in B}\sum_i^N \sum_k^{K_i} \frac{\beta_k}{(2^{b_k}-1)^2}
\end{equation}
\vspace{-4mm}

We jointly address the two objectives in Eqn.~\ref{eq:time} and Eqn.~\ref{eq:variance}, which formulates a bi-objective optimization problem. We apply the weighted sum method to scalarize the objectives~\cite{marler2004survey} and add auxiliary variables to convert it to a pure minimization problem:

\vspace{-7mm}
\begin{equation}\label{eq:weight_sum}
\small
    \begin{aligned}
    & \min_{b_k \in B} \lambda \sum_i^N \sum_k^{K_i} \frac{\beta_k}{(2^{b_k}-1)^2} + (1-\lambda) Z,\ \lambda \in [0, 1]\\
    & s.t._{1 \leq i \leq N} \ \ \theta_i \sum_k^{K_i} D_k^l b_k + \gamma_i \leq Z,\ Z > 0
    \end{aligned} 
\end{equation}
\vspace{-6mm}

We convert the problem to a Mixed Integer Linear Program by viewing it as an Assignment Problem in the combinatorial optimization field and use off-the-shelf solvers (e.g., GUROBI~\cite{gurobi}) to obtain the bit-width assignments. 
To better adapt to the value change of some parameters in $\beta_k$ (e.g., the minimum and maximum values in message vectors) in training, we periodically re-solve the problem (Sec.~\ref{sec:quant_adaptive}). 
To reduce the overhead for solving the optimization, 
for one layer's forward or backward pass, we order messages transferred in each device pair according to their $\beta_k$ values, and then divide them into groups to reduce the number of variables; messages in a group share the same bit-width assignment. We empirically set the size of message groups, which can be found in Appendix~\ref{app:config}.
\section{Evaluation}\label{sec:exp}





\begin{table}[!t]
\caption{Graph datasets in our experiments. 
}
\label{tab:dataset}
\vskip 0.05in
\resizebox{\linewidth}{!}{
\begin{tabular}{cccccc}
    \toprule
    \textbf{Dataset} & \textbf{\#Nodes} & \textbf{\#Edges} & \textbf{\#Features} & \textbf{\#Classes} & \textbf{Size} \\
    \midrule
     Reddit & 232,965 & 114,615,892 & 602 & 41 & 3.53GB\\
     Yelp~\cite{Zeng2020GraphSAINTGS} & 716,847 & 6,977,410 & 300 & 100 & 2.10GB\\
     ogbn-products & 2,449,029 & 61,859,140 & 100 & 47 & 1.38GB  \\
     AmazonProducts & 1,569,960 & 264,339,468 & 200 & 107 & 2.40GB\\
    \bottomrule
\end{tabular}}
\vspace{-3mm}
\end{table}

\textbf{Implementation.} We build AdaQP on top of DGL 0.9~\cite{wang2019deep} and PyTorch 1.11~\cite{paszke2019pytorch}, leveraging DGL for graph-related data storage and operations and using PyTorch distributed package for process group's initialization and communication. Before training begins, DGL's built-in METIS algorithm partitions the graph; each training process is wrapped to only one device (GPU) and broadcasts the remote node indices (built from DGL's partition book) to create sending and receiving node index sets for fetching messages from others. To support computation-communication parallelization, we integrate our customized  \textit{distributed graph aggregation layer} into PyTorch's Autograd mechanism. 
To support extremely low bit-width message compression, we follow~\cite{liu2021exact} to merge all the quantized messages with lower precision (4-bit or 2-bit) into uniform 8-bit byte streams. To transfer multiple bit-width quantized messages and match them with corresponding buffers, we first group messages according to their assigned bit-width, perform single bit-width quantization to each group and then concatenate all groups into a byte array for transmission. After communication, each training process recovers full-precision messages from the byte array based on a bit-retrieval index set maintained and updated by the Adaptive Bit-width Assigner.

\textbf{Experimental Settings.} We evaluate AdaQP on four large benchmark datasets~\cite{hamilton2017inductive, Zeng2020GraphSAINTGS, hu2020open}, detailed in Table~\ref{tab:dataset}. The transductive graph learning task on Reddit and ogbn-products is single-label node classification, while the multi-label classification task is performed on Yelp and AmazonProducts. We use accuracy and F1-score (micro) as the 
model performance metric for these two tasks, respectively, and 
refer to them both as {\em accuracy}. All datasets follow the ``fixed-partition'' splits 
with METIS. We train two representative models, GCN~\cite{Kipf2017SemiSupervisedCW} and full-batch GraphSAGE~\cite{hamilton2017inductive}. To ensure a fair comparison, we unify all the model-related and training-related hyper-parameters throughout all experiments, whose details can be found in Appendix~\ref{app:config}. The model layer size and the hidden layers' dimensions are set to 3 and 256, respectively, and the learning rate is fixed at 0.01.
Experiments are conducted on two servers (Ubuntu 20.04 LTS) connected by 100Gbps Ethernet, each having two Intel Xeon Gold 6230 2.1GHz CPUs, 256GB RAM and four NVIDIA Tesla V100 SXM2 32GB GPUs.


\begin{table*}[!t]
\caption{Training performance comparison among AdaQP and other works. 
The best \textbf{accuracy} 
and \textbf{training throughput} in each set of experiments are marked in bold.}
\label{tab:main}
\vskip 0.1in
\begin{small}
\resizebox{\textwidth}{!}{
\begin{tabular}{c|c|cccc|c|c|cccc}
    \toprule
    \textbf{Dataset} & \textbf{Partitions} & \textbf{Model} & \textbf{Method} & \textbf{Accuracy(\%)} & \textbf{Throughput (epoch/s)} & \textbf{Dataset} & \textbf{Partitions} & \textbf{Model} & \textbf{Method} & \textbf{Accuracy(\%)} & \textbf{Throughput (epoch/s)} \\
    \midrule
    \multirow{16}{*}{Reddit}& \multirow{8}{*}{2M-1D}& \multirow{4}{*}{GCN} & Vanilla & \textbf{95.36±0.03} & 0.99 & \multirow{16}{*}{Yelp}& \multirow{8}{*}{2M-1D}& \multirow{4}{*}{GCN} & Vanilla & \textbf{44.24±0.19} & 1.18\\
    & & & PipeGCN & $\dagger$ & $\dagger$ & & & & PipeGCN & $\dagger$ &$\dagger$ \\
   & & & SANCUS & 94.73±0.17 & 1.11 (1.12$\times$)& & & & SANCUS & 20.75±2.44&0.80 \\
   & & & AdaQP & \textbf{95.36±0.02} & \textbf{2.17 (2.19$\times$)} & & & & AdaQP & 43.96±0.15 & \textbf{3.04 (2.58$\times$)}\\
   \cline{3-6}
   \cline{9-12}
   & & \multirow{4}{*}{GraphSAGE} & Vanilla & 96.50±0.03 & 0.94 & & & \multirow{4}{*}{GraphSAGE} & Vanilla & 64.65±0.08 & 1.11\\
   & & & PipeGCN & \textbf{96.62±0.00} & \textbf{3.72 (3.96$\times$)} & & & & PipeGCN & 63.88±0.06 & 2.63 (2.37$\times$)\\
   & & & SANCUS & $\dagger$ & $\dagger$ & & & & SANCUS & $\dagger$ &$\dagger$ \\
   & & & AdaQP & 96.49±0.02 & 2.13 (2.27$\times$) & & & & AdaQP & \textbf{64.72±0.13} & \textbf{3.15 (2.83$\times$)}\\
   \cline{2-6}
   \cline{8-12}
   & \multirow{8}{*}{2M-2D} & \multirow{4}{*}{GCN} & Vanilla & 95.35±0.04 & 1.13 & & \multirow{8}{*}{2M-2D} & \multirow{4}{*}{GCN} & Vanilla & \textbf{43.86±0.62} & 1.57\\
   & & & PipeGCN & $\dagger$ & $\dagger$ & & & & PipeGCN & $\dagger$ &$\dagger$ \\
   & & & SANCUS & 94.90±0.02 & 1.48 (1.31$\times$)& & & & SANCUS & 20.78±0.2.45 & 0.66\\
   & & & AdaQP & \textbf{95.38±0.03} & \textbf{2.65 (2.35$\times$)} & & & & AdaQP & 43.84±0.63 & \textbf{3.64 (2.32$\times$)}\\
   \cline{3-6}
   \cline{9-12}
   & & \multirow{4}{*}{GraphSAGE} & Vanilla & 96.55±0.03 & 1.16 & & & \multirow{4}{*}{GraphSAGE} & Vanilla & 64.67±0.12 & 1.19\\
   & & & PipeGCN & \textbf{96.67±0.01} & \textbf{3.13 (2.70$\times$)}& & & & PipeGCN & 63.73±0.14 & 2.32 (1.95$\times$)\\
   & & & SANCUS & $\dagger$ & $\dagger$ & & & & SANCUS & $\dagger$ &$\dagger$ \\
   & & & AdaQP & 96.53 ± 0.04 &2.65 (2.28$\times$)& & & & AdaQP & \textbf{64.78±0.05}& \textbf{3.58 (3.01$\times$)}\\
   \cline{1-6}
   \cline{7-12}
    \multirow{16}{*}{ogbn-products}& \multirow{8}{*}{2M-2D}& \multirow{4}{*}{GCN} & Vanilla & 75.14±0.41 & 0.61 & \multirow{16}{*}{AmazonProducts}& \multirow{8}{*}{2M-2D}& \multirow{4}{*}{GCN} & Vanilla & 51.45±0.12 & 0.42\\
    & & & PipeGCN & $\dagger$ & $\dagger$ & & & & PipeGCN $\dagger$ &$\dagger$\\
   & & & SANCUS & 71.52±0.13 & 0.26 & & & & SANCUS & 20.83±0.18& 0.32\\
   & & & AdaQP & \textbf{75.32±0.28} & \textbf{1.65 (2.70$\times$)}& & & & AdaQP & \textbf{51.50±0.08} & \textbf{1.16 (2.76$\times$)}\\
   \cline{3-6}
   \cline{9-12}
   & & \multirow{4}{*}{GraphSAGE} & Vanilla & \textbf{78.90±0.17} & 0.63& & & \multirow{4}{*}{GraphSAGE} & Vanilla & \textbf{75.69±1.32} & 0.46\\
   & & & PipeGCN & 77.82±0.01 & 1.10 (1.75$\times$)& & & & PipeGCN & 71.96±0.00 & 0.99 (2.15$\times$)\\
   & & & SANCUS & $\dagger$ & $\dagger$ & & & & SANCUS & $\dagger$ &$\dagger$ \\
   & & & AdaQP & 78.85±0.20 & \textbf{1.67 (2.65$\times$)}& & & & AdaQP & \textbf{75.69 ± 1.33} & \textbf{1.21 (2.63$\times$)}\\
   \cline{2-6}
   \cline{8-12}
   & \multirow{8}{*}{2M-4D} & \multirow{4}{*}{GCN} & Vanilla & 75.11±0.09 & 0.79& & \multirow{8}{*}{2M-4D} & \multirow{4}{*}{GCN} & Vanilla & 51.38±0.16 & 0.58\\
   & & & PipeGCN & $\dagger$ & $\dagger$ & & & & PipeGCN & $\dagger$ &$\dagger$ \\
   & & & SANCUS & 71.99±0.16 & 0.21 & & & & SANCUS & 21.22±0.07 & 0.27\\
   & & & AdaQP & \textbf{75.30±0.17} & \textbf{2.18 (2.76$\times$)}& & & & AdaQP & \textbf{51.56±0.20} & \textbf{1.60 (2.76$\times$)}\\
   \cline{3-6}
   \cline{9-12}
   & & \multirow{4}{*}{GraphSAGE} & Vanilla & 78.89±0.09 & 0.77 & & & \multirow{4}{*}{GraphSAGE} & Vanilla & 75.80±1.16 & 0.62\\
   & & & PipeGCN & 76.67±0.01& 1.10 (1.43$\times$)& & & & PipeGCN & 71.91±0.00 & 1.02 (1.65$\times$)\\
   & & & SANCUS & $\dagger$ & $\dagger$ & & & & SANCUS & $\dagger$ &$\dagger$ \\
   & & & AdaQP & \textbf{78.90±0.08} & \textbf{2.15 ( 2.79$\times$)}& & & & AdaQP & \textbf{75.98±1.18}& \textbf{1.61 (2.60$\times$)}\\
    \bottomrule
\end{tabular}}
\end{small}
\vskip -0.05in
\end{table*}

\subsection{Expediting Training While Maintaining Accuracy}
\label{sec:exp_main}
First, we show that AdaQP can drastically improve the training throughput while still obtaining high accuracy. We compare its performance with Vanilla and two other SOTA methods: PipeGCN~\cite{wan2022pipegcn} and SANCUS~\cite{peng2022sancus}, both of which show significant advantages over previous works~\cite{Jia2020ImprovingTA, Cai2021DGCLAE, tripathy2020reducing, Thorpe2021DorylusAS}. We implement Vanilla ourselves and use the open-source code of the other two to reproduce all the results. Note that PipeGCN only implements GraphSAGE while SANCUS implements GCN, so we only show their results on their respective supported GNNs. We run training of each model independently for three times, and the average and standard deviation are presented in Table~\ref{tab:main}. All the methods use the same number of training epochs.

We observe that AdaQP achieves the highest training speed and the best accuracy in the 14/16 and 12/16 sets of experiments,
respectively. Specifically, AdaQP is 2.19 $\sim$ 3.01$\times$ faster than Vanilla with at most 0.30\% accuracy degradation. By carefully and adaptively assigning bit-widths for messages, AdaQP can even improve accuracy up to 0.19\%. Compared to AdaQP, PipeGCN and SANCUS not only are slower in most settings but also introduce intolerable model accuracy loss. This is because AdaQP does not rely on stale messages that slow down training convergence.
 What is more, properly applied quantization can benefit training due to the regularization effect of quantization noise introduced to model weights~\cite{courbariaux2015binaryconnect}. 

Note that PipeGCN achieves higher training throughput than AdaQP on Reddit, which is because Reddit is much denser than others. This nature helps the cross-iteration pipeline design of PipeGCN (hiding communication overheads in computation) but does not always hold for other graphs. AdaQP does not rely on this prior property of graphs and can obtain consistent acceleration on distributed full-graph training. We also notice that SANCUS's performance is even worse than Vanilla's under many settings. This is because it adopts sequential node broadcasts, which is less efficient than the ring all2all communication pattern adopted by Vanilla. 

\begin{figure*}[t]
  \centering
  \includegraphics[width=0.9\textwidth, height=0.3\linewidth]{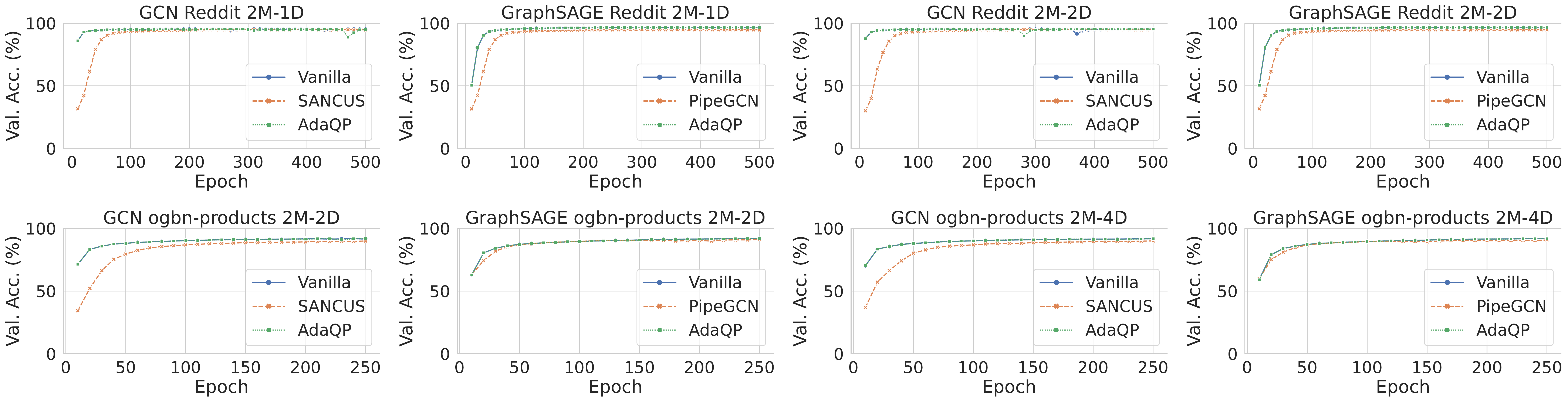}
  \vspace{-5mm}
  \caption{Epoch to validation accuracy comparison among vanilla full-graph training, PipeGCN, SANCUS and AdaQP. 
  }
  \vspace{-5mm}
  \label{fig:conver_rate}
\end{figure*}

\subsection{Preserving Convergence Rate}
\label{sec:exp_conver}
To verify the theoretical analysis in Sec.~\ref{sec:theory} that AdaQP is able to maintain the same training convergence rate as Vanilla ($O(T^{-1})$), we show the training curves 
of all methods on Reddit and ogbn-products in Fig.~\ref{fig:conver_rate} (the complete comparison can be found in Appendix~\ref{app:add_exp}). We observe that our training curves almost coincide with those of Vanilla, verifying the theoretical training convergence guarantee in Sec.~\ref{sec:conver}. On the other hand, both PipeGCN and SANCUS lead to slower training convergence, which is also consistent with their theoretical analysis (meaning that more training epochs will be needed if they intend to achieve the same accuracy as vanilla full-graph training and AdaQP). 
To further illustrate the end-to-end training expedition gains of AdaQP, we show the wall-clock time (the total training time, and for AdaQP, wall-clock time contains both the bit-width assignment time and the actual training time) of training among all the methods on AmazonProducts in Table~\ref{tab:wall_clock}. The complete comparison can be found in Appendix~\ref{app:add_exp}.

\begin{table}[!t]
    \centering
    \caption{Training wall-clock time comparison between AdaQP and other methods on AmazonProducts. The best \textbf{wall-clock time} is marked in bold.}
    \vskip 0.15in
    \label{tab:wall_clock}
    \begin{small}
        \resizebox{\linewidth}{!}{
\begin{tabular}{c|c|ccc}
    \toprule
    \textbf{Dataset} & \textbf{Partitions} & \textbf{Model} & \textbf{Method} & \textbf{Wall-clock Time (s)}\\
    \midrule
    \multirow{16}{*}{AmazonProducts}& \multirow{8}{*}{2M-2D}& \multirow{4}{*}{GCN} & Vanilla & 2874.77 \\
    & & & PipeGCN & $\dagger$ \\
    & & & SANCUS & 3782.44 \\
    & & & AdaQP & \textbf{1053.51}\\
   \cline{3-5}
    & & \multirow{4}{*}{GraphSAGE} & Vanilla & 2597.21 \\
    & & & PipeGCN & 1212.65 \\
    & & & SANCUS & $\dagger$ \\
    & & & AdaQP & \textbf{1008.34} \\
   \cline{3-5}
   & \multirow{8}{*}{2M-4D} & \multirow{4}{*}{GCN} & Vanilla & 2057.70 \\
   & & & PipeGCN & $\dagger$ \\
   & & & SANCUS & 3880.68 \\
   & & & AdaQP & \textbf{806.29}\\
   \cline{3-5}
   & & \multirow{4}{*}{GraphSAGE} & Vanilla & 1927.85 \\
   & & & PipeGCN & 1171.38 \\
   & & & SANCUS & $\dagger$ \\
   & & & AdaQP & \textbf{771.52}\\
    \bottomrule
\end{tabular}
}
\end{small}
\vspace{-5mm}
\end{table}

\subsection{Striking Better Trade-off with Adaptive Message Quantization}

We compare our adaptive message quantization scheme 
with the uniform bit-width sampling scheme, which samples a bit-width from $\{2,4,8\}$ 
for each message uniformly and randomly. 
From Table~\ref{tab:algo_compare}, we see that adaptive message quantization obtains higher accuracy with faster training speed in almost all settings. By solving the bi-objective problem, adaptive message quantization can control the overall gradient variance to a certain level while taking into account the training speed, and alleviate stragglers in all communication rounds. However, uniform bit-width sampling is not as robust. In some cases, it leads to apparent accuracy degradation, e.g., 75.03\% vs. 75.32\%. This is because simply performing uniform bit-width sampling can easily assign lower bit-widths (2 or 4) to messages with large $\beta$ values, thus introducing significant variance (Sec.~\ref{sec:theory}) to model gradients and hurting the accuracy.

\begin{table}[!t]
\caption{Accuracy comparison between uniform bit-width sampling and adaptive message quantization on ogbn-products. The \textbf{best accuracy} is marked in bold. 
}
\label{tab:algo_compare}
\vskip 0.15in
\begin{small}
\resizebox{\linewidth}{!}{
\begin{tabular}{ccccc}
    \toprule
    \textbf{Partitions} & \textbf{Model} & \textbf{Method} & \textbf{Accuracy (\%)} & Throughput (epoch/s)\\
    \midrule
     \multirow{4}{*}{2M-2D} & \multirow{2}{*}{GCN} & Uniform & 75.03±0.36 & 1.70\\
     &  & Adaptive & \textbf{75.32±0.28} & 1.65 \\
     & \multirow{2}{*}{GraphSAGE} & Uniform & 78.84±0.23 & 1.64\\
     &  & Adaptive & \textbf{78.85±0.20} & 1.67\\
     \multirow{4}{*}{2M-4D} & \multirow{2}{*}{GCN} & Uniform & 75.16±0.16 & 2.14\\
     &  & Adaptive & \textbf{75.30±0.17} & 2.18\\
     & \multirow{2}{*}{GraphSAGE} & Uniform & 78.85±0.08 & 2.07\\
     &  & Adaptive & \textbf{78.90±0.08} & 2.15\\    
    \bottomrule
\end{tabular}}
\end{small}
\vspace{-3mm}
\end{table}

\subsection{Time Breakdown
}
\label{sec:exp_break}

\begin{figure}[!t]
    \centering
    \subfigure[Per-epoch time]{
    \includegraphics[width=\linewidth]{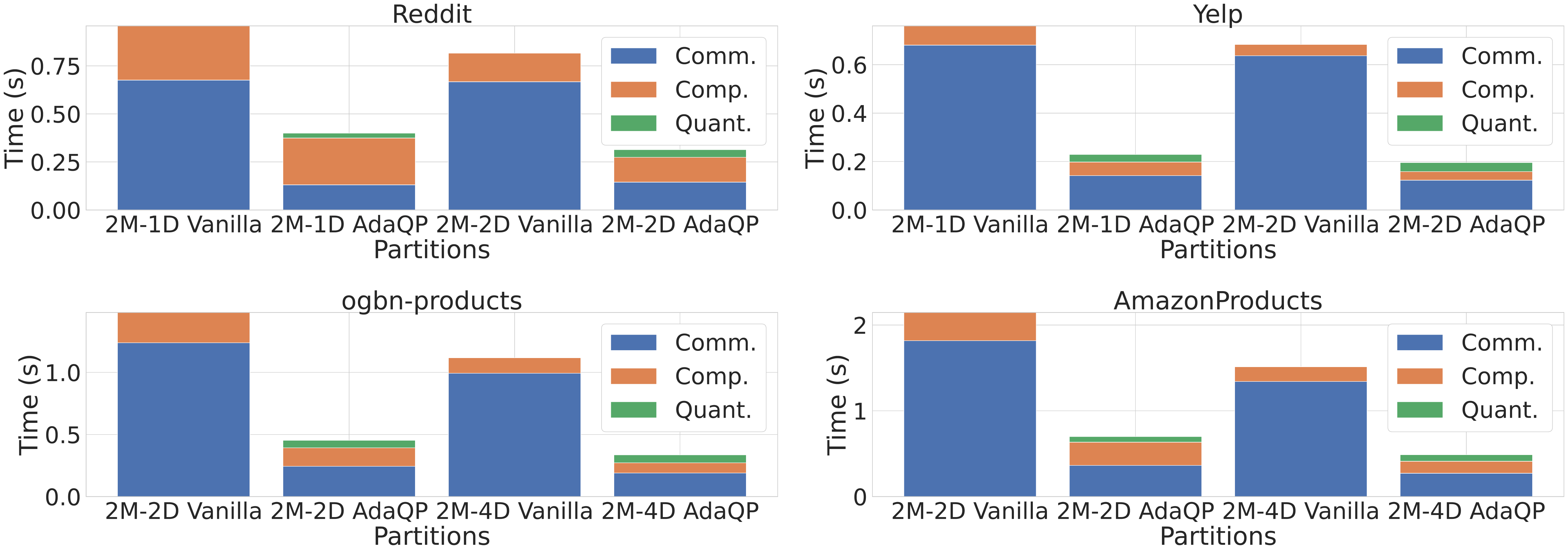}
    \label{fig:exp_iter_break}
    }
    \subfigure[Wall-clock time]{
    \includegraphics[width=\linewidth]{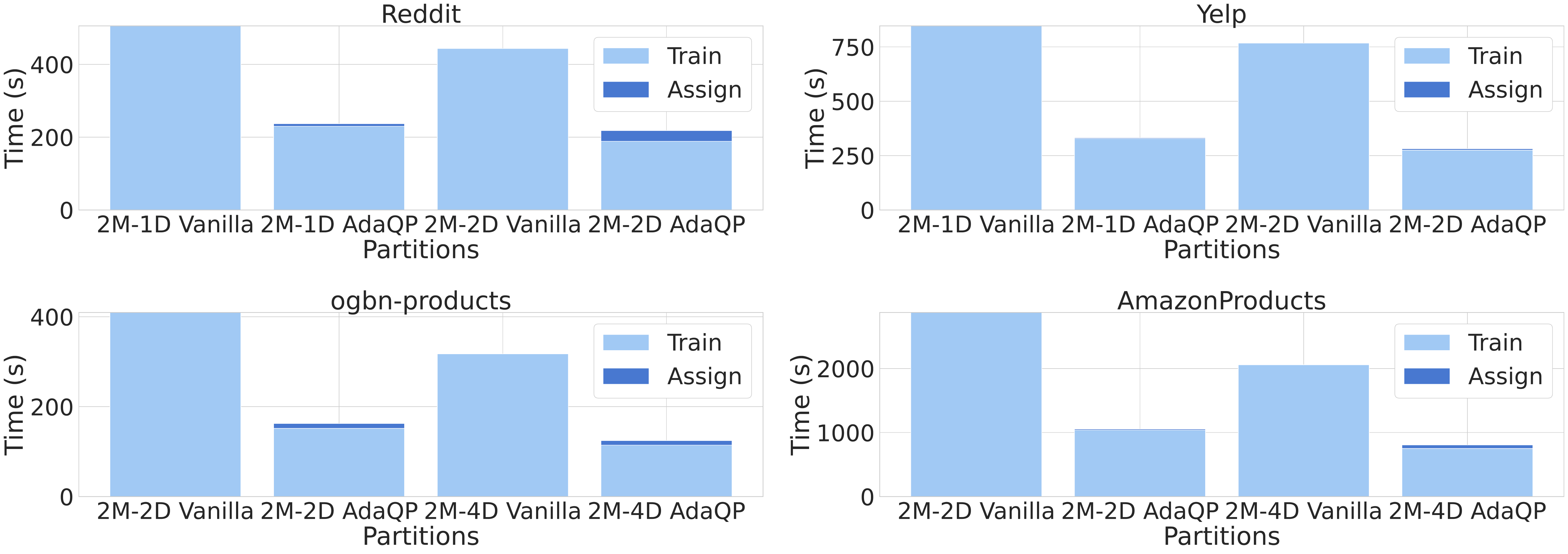}
    \label{fig:exp_wall_break}
    }
    \vspace{-4mm}
    \caption{Time breakdown of Vanilla and AdaQP. 
    {\em quant.} denotes the sum of quantization and de-quantization times.
    }
    \label{fig:breakdown}
    \vspace{-4mm}
\end{figure}

To understand the exact training throughput improvement and the extra overheads brought by AdaQP, we break down per-epoch training time into three parts (communication, computation and quantization time) and the wall-clock time into two parts (assignment time and actual training time). 
We provide the results of training GCN on all datasets in Fig.~\ref{fig:breakdown}. For AdaQP, computation time only includes the marginal graph's computation time since that of the central graph is hidden within communication time (Sec.~\ref{sec:2overhead}). Fig.~\ref{fig:exp_iter_break} shows that compared to communication and computation time reduction benefits brought by AdaQP, the extra quantization cost is almost negligible. Specifically, for AdaQP, the overall quantization overheads are only 5.53\% $\sim$ 13.88\% of per-epoch training time, while the reductions in communication time and computation time are 78.29\% $\sim$ 80.94\% and 13.16\% $\sim$ 39.11\%, respectively. Similar observations can be made in Fig.~\ref{fig:exp_wall_break}, where the average time overhead for bit-width assignment is 5.43\% of the wall-clock time. 

\subsection{Senstivity Analysis}
\label{sec:exp_sen}

\begin{figure}[!t]
  \centering
  \includegraphics[width=\linewidth]{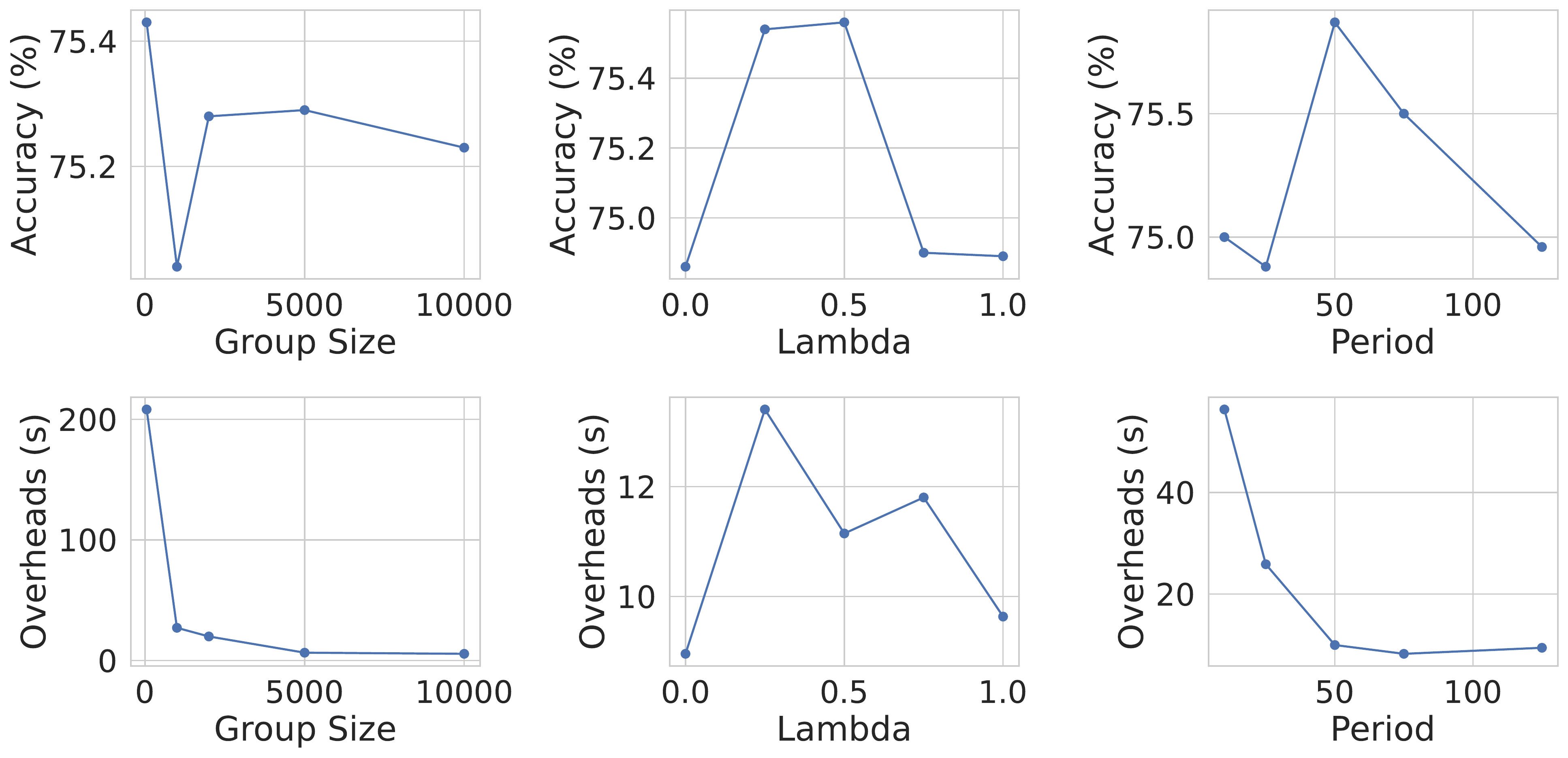}
  \vspace{-5mm}
  \caption{Sensitivity experiments on group size, $\lambda$, and re-assignment period. 
  }
  \label{fig:exp_senstivity}
  \vspace{-5mm}
\end{figure}

There are three hyper-parameters that determine the performance and overhead of adaptive message quantization in AdaQP: a) \textit{group size} 
of messages, which determines the number of variables in Problem~\ref{eq:weight_sum}; b) $\lambda$, which decides the relative weight between time objective and variance objective in the bi-objective minimization problem; c) bit-width re-assignment \textit{period}, which influences the total assignment overhead and the amount of traced data. We perform sensitivity experiments on these hyper-parameters by training GCN on 2M-4D partitioned ogbn-products (since this setting shows the largest accuracy gap between Vanilla and AdaQP). As shown in Fig.~\ref{fig:exp_senstivity}, for group size, the highest accuracy is obtained when it adopts the smallest value (50), which also brings much larger assignment overheads. 
As for $\lambda$, setting it to 0 or 1 both degrades the original problem to a single-objective problem, 
the best model accuracy is not achieved in these cases. As mentioned in Sec.~\ref{sec:exp_main}, quantization can serve as a form of regularization; just rendering the lowest quantization variance ($\lambda=1$) or just pursuing the highest throughput regarding the variance ($\lambda=0$) is not the best choice to fully utilize the regularization effect of quantization. For the re-assignment period, a moderate period length (50) leads to the best model accuracy. How to automatically decide the best values for these hyper-parameters warantees further investigation, e.g., using a reinforcement learning method or searching for the best hyper-parameter combinations. 

\subsection{Scalability of AdaQP}
\label{sec:exp_scale}

\begin{table}[!t]
\begin{small}
\caption{Training throughput on the 6M-4D partition.}
\label{tab:scale}
\vskip 0.05in
\resizebox{\linewidth}{!}{
\begin{tabular}{ccc}
    \toprule
    \textbf{Dataset} & \textbf{Method} & \textbf{Throughput (epoch/s)} \\
    \midrule
     \multirow{2}{*}{ogbn-products} & Vanilla & 0.91\\
     & AdaQP & \textbf{1.63 (1.79 $\times$)}\\
     \multirow{2}{*}{AmazonProducts} & Vanilla & 0.62\\
     & AdaQP & \textbf{1.45 (2.34 $\times$)}\\
    \bottomrule
\end{tabular}}
\vspace{-8mm}
\end{small}
\end{table}

We further evaluate AdaQP's training throughput on 6 machines connected by 100Gps Ethernet (two each have four NVIDIA Tesla V100 SXM2
32GB GPUs and four each have four NVIDIA Tesla A100 SXM4 40GB GPUs). We partition ogbn-products and AmazonProducts among the 24 devices and train GraphSAGE on them. Table~\ref{tab:scale} shows that AdaQP still achieves considerable throughput improvement in this 6M-4D setting, which validates its scalability.

\section{Conclusion}\label{sec:cls}

We propose AdaQP, an efficient system for distributed full-graph GNN training. We are the first to reduce the substantial communication overhead with stochastic integer quantization. We further decompose the local graph partition residing on each device into a central graph and a marginal graph and perform computation-communication parallelization between the central graph's computation and the marginal graph's communication. We provide theoretical analysis to prove that AdaQP achieves similar training convergence rate as vanilla distributed full-graph training, and propose a periodically adaptive bit-width assignment scheme to strike a good trade-off between training convergence and efficiency with negligible extra overheads. Extensive experiments validate the advantages of AdaQP over Vanilla and SOTA works on distributed full-graph training.
\section*{Acknowledgements}
This work was supported in part by grants from Hong
Kong RGC under the contracts HKU 17208920, 17207621 and 17203522.
\bibliography{paper}
\bibliographystyle{mlsys2023}

\appendix
\section{proof}
\label{app:proof}

\subsection{Theorem 1}
\begin{proof}
for any message vector $h_v^l$, the $round_{st}(\cdot)$ operation one of its elements $h_{v,i}^l$ is:

\begin{equation}
\small
round_{st}(h_{v,i}^l)=
\begin{cases}
    \lceil h_{v,i}^l \rceil & p=h_{v,i}^l - \lfloor h_{v,i}^l \rfloor \\
    \lfloor h_{v,i}^l \rfloor & p=1 - (h_{v,i}^l - \lfloor h_{v,i}^l \rfloor) \\
    \end{cases}
\end{equation}

where $p$ is the probability of rounding $h_{v,i}^l$ to certain value. $\lceil \cdot \rceil$ and $\lfloor \cdot \rfloor$ are ceil and floor operations respectively. Since $\lceil h_{v,i}^l \rceil - \lfloor h_{v,i}^l \rfloor=1$, we have $\mathbb{E}[round_{st}(h_{v,i}^l)]=\lceil h_{v,i}^l \rceil (h_{v,i}^l - \lfloor h_{v,i}^l \rfloor)+\lfloor h_{v,i}^l \rfloor (1 - (h_{v,i}^l - \lfloor h_{v,i}^l \rfloor))=h_{v,i}^l$. Therefore, after $\tilde{q}_b(\cdot)$ (Eq.\ref{eq:quant}) and $dq_b(\cdot)$ (Eq.~\ref{eq:dequant}) operations, the expectation of $\hat{h}_v^l$ is:

\begin{equation}
\small
\begin{aligned}
&\mathbb{E}[\hat{h}_v^l]=\mathbb{E}[round_{st}(\frac{h_v^l-Z_v^l}{S_{v_b}^l})S_{v_b}^l + Z_v^l] \\ &=S_{v_b}^l\mathbb{E}[round_{st}(\frac{h_v^l-Z_v^l}{S_{v_b}^l})] + Z_v^l \\
&=h_v^l
\end{aligned}
\end{equation}

as for variance of $\hat{h_v^l}$, we have:

\begin{equation}
\small
    \mathbb{V}ar[\hat{h}_v^l]=(S_{v_b}^l)^2\mathbb{V}ar[round_{st}(\frac{h_v^l-Z_v^l}{S_{v_b}^l})]
\end{equation}
let $h=\frac{h_v^l-Z_v^l}{S_{v_b}^l}$, since $\mathbb{V}ar[h]=\mathbb{E}[h^T h]-\mathbb{E}[h^T]\mathbb{E}[h]$, hence:
\begin{equation}
\begin{aligned}
    &\mathbb{V}ar[round_{st}(\frac{h_v^l-Z_v^l}{S_{v_b}^l})] = \sum_{i}^{D_v^l}\lceil h_i \rceil^2 (h_i - \lfloor h_i \rfloor) \\
    &+ \lfloor h_i \rfloor^2 (1-h_i+\lfloor h_i \rfloor) - h_i^2 \\
    &= \sum_{i}^{D_v^l}(2\lfloor h_i \rfloor h_i + h_i - \lfloor h_i \rfloor^2 - \lfloor h_i \rfloor - h_i^2)
\end{aligned}
\end{equation}

we make the assumption that $\forall i, h_i - \lfloor h_i \rfloor = \sigma \sim $ Uniform(0,1), then $Var[round_{st}(\frac{h_v^l-Z_v^l}{S_{v_b}^l})]=\sum_{i}^{D_v^l}(\sigma - \sigma^2)=\frac{D_v^l}{6}$. Finally, the variance of $\hat{h}_v^l$ is:

\begin{equation}
\small
    \mathbb{V}ar[\hat{h}_v^l]=\frac{D_v^l(S_{v_b}^l)^2}{6}
\end{equation}
\end{proof}

\subsection{Theorem 2}
\begin{proof}
We perform Taylor expansion for $\mathcal{L}(\mathbf{w}_{t+1})$ with  Lagrangian Remainder:

\begin{equation}
\small
\begin{aligned}
& f(\mathbf{w}_{t+1}) = \mathcal{L}(\mathbf{w}_t-\alpha \mathbf{g}_t + \alpha \mathbf{g}_t - \mathbf{\tilde{g}}_t) \\
& = \mathcal{L}(\mathbf{w}_t-\alpha \mathbf{g}_t) + \alpha(\mathbf{g}_t - \mathbf{\tilde{g}}_t)^T \nabla \mathcal{L}(\mathbf{w}_t - \alpha \mathbf{g}_t) \\
& + \frac{1}{2}\alpha^2 (\mathbf{g}_t - \mathbf{\tilde{g}}_t)^T \nabla^2 \mathcal{L}(\mathbf{\epsilon}_t)(\mathbf{g}_t - \mathbf{\tilde{g}}_t)
\end{aligned}
\end{equation}

since $\mathbb{E}[\mathbf{\tilde{g}}_t] = \mathbf{g}_t$ (from Assumption~\ref{assum:conver}), we have:

\begin{equation}\label{Eq:first}
\small
\begin{aligned}
& \mathbb{E}[\mathcal{L}(\mathbf{w}_{t+1})]  \leq \mathbb{E}[\mathcal{L}(\mathbf{w}_t-\alpha \mathbf{g}_t) \\
& + \alpha(\mathbf{g}_t - \mathbf{\tilde{g}}_t)^T \nabla \mathcal{L}(\mathbf{w}_t - \alpha \mathbf{g}_t) \\
& + \frac{1}{2}\alpha^2 L_2||\mathbf{g}_t - \mathbf{\tilde{g}}_t||^2] \\
& \leq \mathbb{E}[\mathcal{L}(\mathbf{w}_t-\alpha \mathbf{g}_t)] + \frac{1}{2}\alpha^2L_2Q^2
\end{aligned}
\end{equation}

where the first inequality is due to the property of Lipschitz continuity, the second inequality is due to the bounded variance property that $\mathbb{E}[||\mathbf{g} - \mathbf{\tilde{g}}||] \leq Q$. We perform similar Taylor Expansion to $\mathcal{L}(\mathbf{w}_t-\alpha \mathbf{g}_t)$ in Eq.~\ref{Eq:first} and take expectation on
both sides:

\begin{equation}\label{eq:second}
\small
    \begin{aligned}
    & \mathbb{E}[\mathcal{L}(\mathbf{w}_t-\alpha \mathbf{g}_t)] = \mathbb{E}[\mathcal{L}(\mathbf{w}_t) - \alpha \mathbf{g}_t^T \nabla \mathcal{L}(\mathbf{w}_t) \\
    & + \frac{1}{2}\alpha^2\mathbf{g}_t^T \nabla \mathcal{L}(\mu_t)\mathbf{g}_t] \\
    & \leq \mathcal{L}(\mathbf{w}_t) - \alpha \mathbb{E}[||\mathbf{g}_t||^2] + \frac{1}{2}\alpha^2 L_2 \mathbb{E}[||\mathbf{g}_t||^2]
    \end{aligned}
\end{equation}

denote $\mathbf{g}_t$ as $\nabla \mathcal{L}(\mathbf{w}_t)$ and plug Eq.~\ref{eq:second} into Eq.~\ref{Eq:first}, we have:
\begin{equation}
    \begin{aligned}
    & (\alpha - \frac{1}{2}\alpha^2 L_2)\mathbb{E}[||\nabla \mathcal{L}(\mathbf{w}_t)||^2] \leq \mathbb{E}[\mathcal{L}(\mathbf{w}_t)] - \mathbb{E}[\mathcal{L}(\mathbf{w}_{t+1})] \\
    & + \frac{1}{2}\alpha^2 L_2 Q^2
    \end{aligned}
\end{equation}
since we assume that this problem exists global minimum (Assumption~\ref{assum:conver}), Summing over t from 1 to T we have:

\begin{equation}
\small
    \begin{aligned}
    \frac{\sum_{t=1}^T\mathbb{E}[||\nabla \mathcal{L}(\mathbf{w}_t)||^2]}{T} \leq \frac{2(\mathcal{L}(\mathbf{w_1})-\mathcal{L}^{\ast})}{T(2\alpha - \alpha^2L_2)} + \frac{\alpha L_2Q^2}{2-\alpha L_2}
    \end{aligned}
\end{equation}

Viewing $t$ as a random variable, we have Theorem~\ref{thm:conver}.

\end{proof}
\subsection{Theorem 3}
\begin{proof}
We denote model gradient matrix in GNNs' layer $l$ with quantization variance as $\frac{\partial \mathcal{L}}{\partial \mathbf{\tilde{W}}^l}$, its full-precision counterpart as $\frac{\partial \mathcal{L}}{\partial \mathbf{W}^l}$, according to the forward pass form in Eq.~\ref{eq:gnn_sum}, we have:

\begin{equation}
\small
\begin{aligned}
    & \frac{\partial \mathcal{L}}{\partial \mathbf{\tilde{W}}^l} =  \sum_v^{|V|}\sigma'(\cdot) \odot \frac{\partial \mathcal{L}}{\partial h_v^l}(\sum_{u}^{\{v\} \cup N(v)} \alpha_{u,v}h_u^{l-1})^T \\
    & = \sum_v^{|V|}\sigma'(\cdot) \odot (\sum_{u}^{\{v\} \cup N_{L}(v)}\alpha_{u,v}\frac{\partial \mathcal{L}}{\partial h_u^l} + \sum_{k}^{N_{R}(v)}\alpha_{k,v}\frac{\partial \mathcal{L}}{\partial \hat{h}_{k_b}^l}) \cdot \\ & (\sum_{u}^{\{v\} \cup N_{L}(v)}\alpha_{u,v}h_u^{l-1} + \sum_{k}^{N_{R}(v)}\alpha_{k,v}\hat{h}_{k_b}^{l-1})^T
\end{aligned}
\end{equation}

Consider that each message quantization operation in the forward and backward pass are independent of each other, we can get the expectation and variance of $\frac{\partial \mathcal{L}}{\partial \mathbf{\tilde{W}}^l}$ based on Theorem~\ref{thm:quant}:

\begin{equation}
\small
    \begin{aligned}
        & \mathbb{E}[\frac{\partial \mathcal{L}}{\partial \mathbf{\tilde{W}}^l}] = \sum_v^{|V|}\sigma'(\cdot) \odot (\sum_{u}^{\{v\} \cup N_{L}(v)}\alpha_{u,v}\frac{\partial \mathcal{L}}{\partial h_u^l} \\
    & + \sum_{k}^{N_{R}(v)}\alpha_{k,v}\mathbb{E}[\frac{\partial \mathcal{L}}{\partial \hat{h}_{k_b}^l}])(\sum_{u}^{\{v\} \cup N_{L}(v)}\alpha_{u,v}h_u^{l-1} \\
    & + \sum_{k}^{N_{R}(v)}\alpha_{k,v}\mathbb{E}[\hat{h}_{k_b}^{l-1}])^T \\
    & = \frac{\partial \mathcal{L}}{\partial \textbf{W}^l}
    \end{aligned}
\end{equation}

for variance, we omit the activation function $\sigma$ since it does not change variance form, we have: 

\begin{equation}
\small
    \begin{aligned}
        & \mathbb{V}ar[\frac{\partial \mathcal{L}}{\partial \mathbf{\tilde{W}}^l}] = \sum_v^{|V|}\mathbb{V}ar[\frac{\partial \mathcal{L}}{\partial h_v^l}(\sum_u^{\{v\} \cup N(v)}\alpha_{u,v}h_u^{l-1})^T] \\
        & = \sum_v^{|V|} \mathbb{E}[(\frac{\partial \mathcal{L}}{\partial h_v^l})^2]\mathbb{E}[(\sum_u^{\{v\} \cup N(v)}\alpha_{u,v}h_u^{l-1})^2]^T \\
        & - \mathbb{E}[\frac{\partial \mathcal{L}}{\partial h_v^l}]^2\mathbb{E}[(\sum_u^{\{v\} \cup N(v)}\alpha_{u,v}h_u^{l-1})^T]^2 \\
        & = \sum_v^{|V|} Var[\frac{\partial \mathcal{L}}{\partial h_v^l}]Var[\sum_u^{\{v\} \cup N(v)}\alpha_{u,v}h_u^{l-1})] \\
        & + \mathbb{V}ar[\frac{\partial \mathcal{L}}{\partial h_v^l}]\mathbb{E}[\sum_u^{\{v\} \cup N(v)}\alpha_{u,v}h_u^{l-1})]^2 \\
        & + \mathbb{V}ar[\sum_u^{\{v\} \cup N(v)}\alpha_{u,v}h_u^{l-1})]\mathbb{E}[\frac{\partial \mathcal{L}}{\partial h_v^l}]^2
    \end{aligned}
\end{equation}

Since the randomness is introduced by $N_R(v)$, utilizing Assumption~\ref{assum:layer} we have:

\begin{equation}
\small
    \begin{aligned}
         & \mathbb{V}ar[\frac{\partial \mathcal{L}}{\partial \mathbf{\tilde{W}}^l}] = \sum_v^{|V|} \mathbb{V}ar[\sum_k^{N_{R}(v)}\alpha_{k,v}\frac{\partial \mathcal{L}}{\partial \hat{h}_{k_b}^l}] \\
         & \mathbb{V}ar[\sum_k^{N_{R}(v)}\alpha_{k,v}h_{k_b}^{l-1}] \\
         & + \mathbb{V}ar[\sum_k^{N_{R}(v)}\alpha_{k,v}\frac{\partial \mathcal{L}}{\partial \hat{h}_{k_b}^l}]\mathbb{E}[\sum_k^{\{v\}\cup N(v)}\alpha_{k,v}h_{k_b}^{l-1}]^2 \\
         & + \mathbb{V}ar[\sum_k^{N_{R}(v)}\alpha_{k,v}h_{k_b}^{l-1}]\mathbb{E}[\sum_k^{\{v\}\cup N(v)}\alpha_{k,v}\frac{\partial \mathcal{L}}{\partial \hat{h}_{k_b}^l}]^2 \\
         & \leq \sum_v^{|V|} \mathbb{V}ar[\sum_k^{N_{R}(v)}\alpha_{k,v}\frac{\partial \mathcal{L}}{\partial \hat{h}_{k_b}^l}] \mathbb{V}ar[\sum_k^{N_{R}(v)}\alpha_{k,v}h_{k_b}^{l-1}] \\
         & + M^2 \mathbb{V}ar[\sum_k^{N_{R}(v)}\alpha_{k,v}\frac{\partial \mathcal{L}}{\partial \hat{h}_{k_b}^l}] + N^2 \mathbb{V}ar[\sum_k^{N_{R}(v)}\alpha_{k,v}h_{k_b}^{l-1}]
    \end{aligned}
\end{equation}

use Theorem~\ref{thm:quant}, we have:
\begin{equation}\label{proof:upper_bound}
\small
    \begin{aligned}
        & \mathbb{V}ar[\frac{\partial \mathcal{L}}{\partial \mathbf{\tilde{W}}^l}] \leq \sum_v^{|V|} (\sum_k^{N_{R}(v)}\alpha^2_{k,v}\frac{D_k^l \cdot (S_{k_b}^l)^2}{6}) \\
         & \cdot (\sum_k^{N_{R}(v)}\alpha^2_{k,v}\frac{D_k^{l-1} \cdot (S_{k_b}^{l-1})^2}{6}) + M^2 \sum_k^{N_R(V)}\alpha^2_{k,v} \frac{D_k^{l}(S_{k_b}^{l})^2}{6} \\
         & + N^2 \sum_k^{N_R(v)}\alpha^2_{k,v} \frac{D_k^{l-1}(S_{k_b}^{l-1})^2}{6})
    \end{aligned}
\end{equation}

We can just let the model gradient matrix's upper bound $Q^l$ in layer $l$ be the gradient variance upper bound in Eqn.~\ref{proof:upper_bound}:
\begin{equation}
\small
    \begin{aligned}
          & Q^l = \sum_v^{|V|}(\sum_{k_1}^{N_R(v)}\sum_{k_2}^{N_R(v)}\alpha^2_{k_1,v}\alpha^2_{k_2,v} \frac{D_{k_1}^{l-1}D_{k_2}^l(S_{k_{1_b}}^{l-1}S_{k_{2_b}}^l)^2}{6}\\
         &+ M^2 \sum_k^{N_R(v)}\alpha^2_{k,v} \frac{D_k^{l}(S_{k_b}^{l})^2}{6}
         + N^2 \sum_k^{N_R(v)}\alpha^2_{k,v} \frac{D_k^{l-1}(S_{k_b}^{l-1})^2}{6})
    \end{aligned}
\end{equation}

\end{proof}

\section{Training Configuration}

We show the training configurations in Table~\ref{tab:config}, where GCN and GraphSAGE share the same configurations. We also include the message group size and the value of $\lambda$ for AdaQP in the table.

\begin{table*}[!b]
\caption{ Training configurations in our experiments.}
\label{tab:config}
\vspace{5mm}
\resizebox{\linewidth}{!}{
\begin{tabular}{cccccccccc}
    \toprule
    \textbf{Dataset} & \textbf{Model Layer} & \textbf{Hidden Dimension} & \textbf{Norm Function} & \textbf{Optimizer} & \textbf{Learning Rate} & \textbf{Dropout} & \textbf{Epoch} & \textbf{Message Group Size} & $\boldsymbol{\lambda}$\\
    \midrule
     Reddit & 3 & 256 & LayerNorm & Adam & 0.01 & 0.5 & 500 & 100 & 0.5 \\
     Yelp & 3 & 256 & LayerNorm& Adam & 0.01 & 0.1 & 1000 & 1000 & 0.5 \\
     ogbn-products & 3 & 256 & LayerNorm & Adam & 0.01 & 0.5 & 250 & 2000 & 0.5 \\
     AmazonProducts & 3 & 256 & LayerNorm & Adam & 0.01 & 0.5 & 1200 & 500 & 0.5 \\
    \bottomrule
\end{tabular}}
\end{table*}

\label{app:config}

\section{Additional Experiments}
\label{app:add_exp}
\subsection{Training Convergence Comparison}
\label{app:add_exp_cinver}

\begin{figure*}[t]
  \centering
  \includegraphics[width=0.8\textwidth, height=0.4\textwidth]{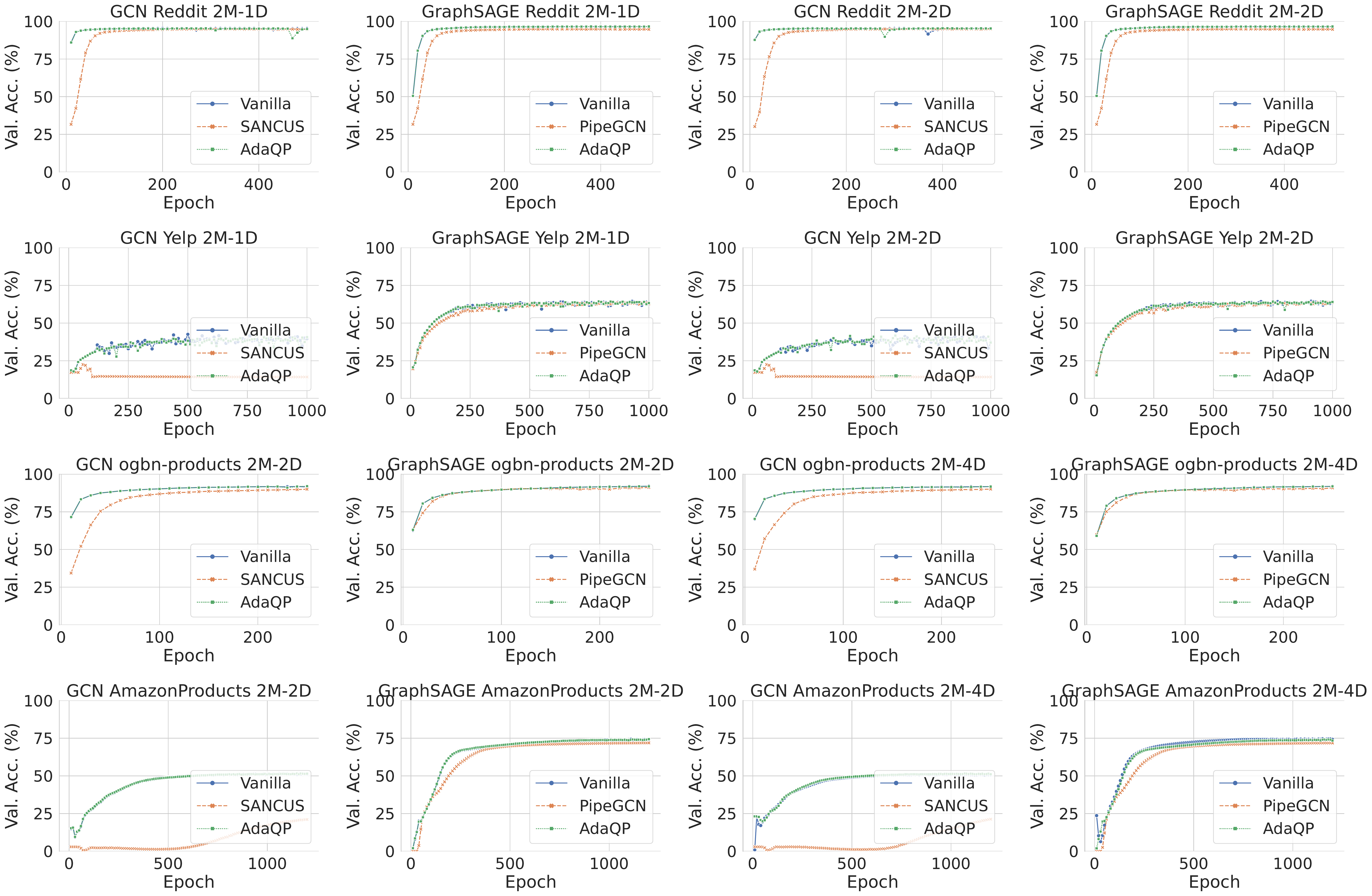}
  \caption{Epoch to validation accuracy comparison among Vanilla, PipeGCN, SANCUS, and AdaQP on all the datasets.}
  \label{fig:full_conver_rate}
  \vskip -0.5in
\end{figure*}

Fig.~\ref{fig:full_conver_rate} shows the training convergence comparison of all the methods on all the datasets under the same experiential settings of Sec.~\ref{sec:exp_main}. As we can see, AdaQP consistently shows a fast convergence rate over other SOTA staleness-based expedition methods, which is similar to the conclusion drawn in Sec.~\ref{sec:exp_conver}.

\subsection{Wall-clock Time Comparison}

We provide the wall-clock time comparison of all the methods on all the datasets in Tab.~\ref{tab:exp_full_wall_clock} under the same experiment settings in Sec.~\ref{sec:exp_main}. For the fairness of comparison, we include the extra bit-width assignment time overheads (details in ~\ref{sec:quant_adaptive}) in the wall-clock time of AdaQP. The results prove that the extra overheads introduced by AdaQP are negligible compared to the overall wall-clock time reduction gains, which is consistent with the time breakdown analysis in Sec~\ref{sec:exp_break}. We still achieve the shortest wall-clock time in 14/16 sets of experiments.

\begin{table*}[htp]
\vspace{-10mm}
\caption{Training wall-clock time comparison between AdaQP and other methods, the best is denoted in bold.}
\label{tab:exp_full_wall_clock}
\vskip 0.15in
\begin{small}
\resizebox{\textwidth}{!}{
\begin{tabular}{c|c|ccc|c|c|ccc}
    \toprule
    \textbf{Dataset} & \textbf{Partitions} & \textbf{Model} & \textbf{Method} & \textbf{Wall-clock Time (s)} & \textbf{Dataset} & \textbf{Partitions} & \textbf{Model} & \textbf{Method} & \textbf{Wall-clock Time (s)} \\
    \midrule
    \multirow{16}{*}{Reddit}& \multirow{8}{*}{2M-1D}& \multirow{4}{*}{GCN} & Vanilla & 505.79& \multirow{16}{*}{Yelp}& \multirow{8}{*}{2M-2D}& \multirow{4}{*}{GCN} & Vanilla & 846.79 \\
    & & & PipeGCN & $\dagger$ & & & & PipeGCN & $\dagger$ \\
   & & & SANCUS & 447.28 & & & & SANCUS & 1249.89\\
   & & & AdaQP & \textbf{237.24} & & & & AdaQP & \textbf{332.37} \\
   \cline{3-5}
   \cline{8-10}
   & & \multirow{4}{*}{GraphSAGE} & Vanilla & 530.46 & & & \multirow{4}{*}{GraphSAGE} & Vanilla & 897.28 \\
   & & & PipeGCN & \textbf{135.29} & & & & PipeGCN & 381.11 \\
   & & & SANCUS & $\dagger$ & & & & SANCUS & $\dagger$ \\
   & & & AdaQP & 246.71 & & & & AdaQP & \textbf{321.54}\\
   \cline{3-5}
   \cline{8-10}
   & \multirow{8}{*}{2M-2D} & \multirow{4}{*}{GCN} & Vanilla & 443.53 & & \multirow{8}{*}{2M-4D} & \multirow{4}{*}{GCN} & Vanilla & 767.47\\
   & & & PipeGCN & $\dagger$ & & & & PipeGCN & $\dagger$ \\
   & & & SANCUS & 335.56 & & & & SANCUS & 1509.41 \\
   & & & AdaQP & \textbf{218.14} & & & & AdaQP & \textbf{281.77}\\
   \cline{3-5}
   \cline{8-10}
   & & \multirow{4}{*}{GraphSAGE} & Vanilla & 429.85 & & & \multirow{4}{*}{GraphSAGE} & Vanilla & 839.96\\
   & & & PipeGCN &  \textbf{159.66} & & & & PipeGCN & 430.63\\
   & & & SANCUS & $\dagger$ & & & & SANCUS & $\dagger$ \\
   & & & AdaQP & 208.34 & & & & AdaQP & \textbf{289.23}\\
   \cline{1-10}
    \multirow{16}{*}{ogbn-products}& \multirow{8}{*}{2M-2D}& \multirow{4}{*}{GCN} & Vanilla & 409.54 & \multirow{16}{*}{AmazonProducts}& \multirow{8}{*}{2M-2D}& \multirow{4}{*}{GCN} & Vanilla & 2874.77 \\
    & & & PipeGCN & $\dagger$ & & & & PipeGCN & $\dagger$ \\
   & & & SANCUS & 940.16 & & & & SANCUS & 3782.44 \\
   & & & AdaQP & \textbf{162.53} & & & & AdaQP & \textbf{1053.51}\\
   \cline{3-5}
   \cline{8-10}
   & & \multirow{4}{*}{GraphSAGE} & Vanilla & 397.91 & & & \multirow{4}{*}{GraphSAGE} & Vanilla & 2597.21 \\
   & & & PipeGCN & 229.11 & & & & PipeGCN & 1212.65 \\
   & & & SANCUS & $\dagger$ & & & & SANCUS & $\dagger$ \\
   & & & AdaQP & \textbf{155.94} & & & & AdaQP & \textbf{1008.34} \\
   \cline{3-5}
   \cline{8-10}
   & \multirow{8}{*}{2M-4D} & \multirow{4}{*}{GCN} & Vanilla & 317.48 & & \multirow{8}{*}{2M-4D} & \multirow{4}{*}{GCN} & Vanilla & 2057.70 \\
   & & & PipeGCN & $\dagger$ & & & & PipeGCN & $\dagger$ \\
   & & & SANCUS & 1186.68 & & & & SANCUS & 3880.68 \\
   & & & AdaQP & \textbf{124.67} & & & & AdaQP & \textbf{806.29}\\
   \cline{3-5}
   \cline{8-10}
   & & \multirow{4}{*}{GraphSAGE} & Vanilla & 326.05 & & & \multirow{4}{*}{GraphSAGE} & Vanilla & 1927.85 \\
   & & & PipeGCN & 229.31 & & & & PipeGCN & 1171.38 \\
   & & & SANCUS & $\dagger$ & & & & SANCUS & $\dagger$ \\
   & & & AdaQP & \textbf{133.93} & & & & AdaQP & \textbf{771.52}\\
    \bottomrule
\end{tabular}}
\end{small}
\vskip -0.1in
\end{table*}



\end{document}